\documentclass{article}

\usepackage[preprint]{neurips_2026}

\usepackage[utf8]{inputenc} 
\usepackage[T1]{fontenc}    
\usepackage{hyperref}       
\usepackage{url}            
\usepackage{booktabs}       
\usepackage{amsfonts}       
\usepackage{nicefrac}       
\usepackage{microtype}      
\usepackage{amsmath}
\usepackage{wrapfig}

\usepackage{multirow}
\usepackage{graphicx}
\usepackage[table]{xcolor}
\usepackage{subcaption}

\newcommand{\sr}[1]{\cellcolor{blue!8}{#1}}

\definecolor{compRed}{HTML}{8B2323}    
\definecolor{compBlue}{HTML}{2B59C3}   
\definecolor{compGreen}{HTML}{4E7D2E}  

\newcommand{\obj}[1]{\textbf{\textcolor{compRed}{#1}}}
\newcommand{\attr}[1]{\textbf{\textcolor{compBlue}{#1}}}
\newcommand{\rel}[1]{\textbf{\textcolor{compGreen}{#1}}}

\newcommand{\name}{COMPASS}

\title{Diagnosing the Sources of Compositional Failure in Vision-Language Models: A Controlled Analysis}

\author{
    Mona Gandhi$^1$,  
  Cenk Merih Olcay$^1$, 
    Kuan-Chieh Lo$^1$, 
    Santiago Castro$^2$, \\
    \textbf{Christopher W. Myers$^1$, 
    Srinivasan Parthasarathy$^1$ }
    \And
    \textbf{$^1$The Ohio State University, 
    $^2$Netflix Research }
    \\
    gandhi.255@osu.edu
}

\begin{document}

\maketitle

\begin{abstract}

    Vision-language models (VLMs) often struggle with compositional reasoning tasks, but the reasons for this underperformance remain unclear. A common hypothesis is that models struggle to integrate multiple components, leading to training interventions to improve compositional binding. However, this assumption has never been directly quantified. Existing benchmarks evaluate captions only in their composed form, making it impossible to separate the cost of joint reasoning from the cost of recognizing individual components under increasing load.
    We introduce \textbf{\name{}} (COMPositional Analysis of SkillS), a controlled evaluation framework designed to isolate and measure the distinct factors underlying compositional failure. By comparing performance on composed captions with their decomposed counterparts under matched perturbations, we directly quantify the cost of compositional integration across \textbf{87K} image-caption pairs. Across multiple VLMs, this gap is real but partial, accounting for only part of the observed degradation. This motivates a finer-grained investigation into what additional factors govern model behavior. We analyze performance at the level of individual skills: object detection, attribute binding, and relation reasoning, using skill-targeted perturbations across \textbf{274K} image-caption pairs. We find a consistent skill-specific pattern: each skill degrades primarily with the count of its own primitive type (self-load), while cross-load effects are predominantly positive, suggesting that primitives of different types provide useful grounding context. This pattern holds across standard contrastive encoders, explicitly trained compositional reasoning models, and non-contrastive architectures. These findings show that compositional degradation reflects multiple separable factors that cannot be reduced to joint reasoning alone. \name{} provides a controlled structure for diagnosing these factors independently, enabling a more precise evaluation of how future models improve along distinct dimensions of compositional reasoning.

\end{abstract}

\section{Introduction}
\label{sec:intro}

Vision-language models (VLMs) are often deployed in settings that demand precise scene understanding, including image-text retrieval~\citep{CLIP, ALIGN}, referring expression grounding~\citep{glip, peng2023kosmos2groundingmultimodallarge}, and visual question answering~\citep{blip, liu2023llava}, among others. In each of these settings, success requires more than recognizing individual objects. It requires understanding how objects relate to their attributes and to one another. \textit{Compositional understanding}, the ability to understand and produce novel combinations of known concepts, is a fundamental principle of human cognition~\citep{partee1984compositionality, bottou2011machinelearningmachinereasoning} that neural networks have long struggled to replicate~\citep{hupkes2020compositionalitydecomposedneuralnetworks}. For VLMs, compositional understanding remains a persistent and open challenge.

\citet{ma2023crepe} show that model performance degrades monotonically as caption complexity increases, frequently nearing random chance at high complexity, regardless of model architecture or training dataset size. Yet despite extensive benchmarking~\citep{aro, hsieh2023sugarcrepe, winoground}, we still lack a clear answer to a basic question: what is actually failing? Most prior work has attributed compositional failure to the difficulty of integrating multiple primitives simultaneously and has focused on improving this by using training objectives, data augmentation, and architectural modifications that facilitate compositional binding~\citep{aro, svlc, glip}. However, existing benchmarks evaluate captions only in their composed form, without isolating the distinct factors that contribute to the observed degradation, leaving the role of joint reasoning unquantified. \textit{To what extent does joint reasoning actually contribute to this degradation?} And if it is only a partial explanation, \textit{what other factors govern model behavior under increasing complexity?}

To directly test these questions, we introduce \textbf{\name{}} (COMPositional Analysis of SkillS). We construct captions from scene graphs, where each caption is composed of primitives --objects, attributes, and relations -- corresponding to three core visual skills: object detection, attribute binding, and relation reasoning. By systematically varying both the type and count of these primitives, \name{} enables two targeted analyses. First, by pairing composed captions with decomposed counterparts under matched perturbations, we isolate the \textbf{compositional integration gap}, the cost attributable to joint reasoning alone. Second, by constructing skill-targeted negatives and modeling performance as a function of per-type primitive counts, we measure \textbf{skill load}, that is, how each skill is affected by each primitive count. We further distinguish between \textbf{self-load}, degradation induced by increasing the count of the same primitive type being tested, and \textbf{cross-load}, degradation induced by increasing the count of other primitive types. In total, we evaluate compositional integration on \textbf{87K} image--caption pairs and skill load on \textbf{274K} pairs across object, attribute, and relation skills.

Across a diverse set of vision-language models spanning standard contrastive encoders, explicitly trained compositional reasoning models, and non-contrastive architectures, we observe that compositional integration incurs a consistent cost but does not fully account for the observed degradation. Performance also varies systematically with primitive counts, with each skill exhibiting sensitivity to its own components and differing behavior across primitive types. These patterns suggest that compositional failure reflects multiple interacting factors. \name{} provides the controlled structure to diagnose these factors independently through two targeted analyses, compositional integration gap and skill load, enabling more precise evaluation of how future models improve along distinct dimensions of compositional generalization. We hope these analyses motivate skill-specific interventions that ensure individual skills remain robust as the primitive load increases.

\section{Related Work}
\paragraph{Vision-language compositionality benchmarks.}
Compositional understanding remains a persistent challenge for VLMs, documented across diverse phenomena: word order and caption matching \citep{winoground}, spatial relations and linguistic grounding \citep{valse, aro}, verb and relation understanding \citep{svo}, and negation comprehension \citep{negbench}. CREPE \citep{ma2023crepe} introduces complexity-aware evaluation, establishing a monotonic degradation curve with respect to entity count that is complexity. SugarCREPE \citep{hsieh2023sugarcrepe} addresses benchmark hackability by using LLM-generated negatives via single-negative retrieval. Despite this progress, existing benchmarks attribute performance degradation to joint reasoning, the cost of simultaneously composing across multiple primitive types (e.g., objects, attributes, and relations) without directly quantifying how much of the observed decline actually stems from joint reasoning. 
COMPASS addresses this limitation by introducing a compositional integration gap that directly measures the cost of joint reasoning under matched experimental conditions.

Several works move toward finer-grained evaluation. VL-CheckList \citep{VL-CheckList} reveals model-specific variability across primitive types, 
and SugarCREPE \citep{hsieh2023sugarcrepe} decomposes negatives by primitive type, showing that relation negatives are harder than attribute and object negatives. However, benchmarks that vary in caption complexity treat it as a single axis, conflating the distinct contributions of individual primitive types.  We address this gap by decomposing complexity into per-primitive axes to directly measure how each primitive type contributes to performance degradation. To understand these effects at the skill level, we construct skill-targeted negatives that perturb one primitive type at a time, isolating the load imposed independently.

\paragraph{Improving compositional reasoning in VLMs.}
A range of interventions has targeted compositional reasoning through training: hard-negative fine-tuning \citep{aro, zhang2024contrastingintramodalrankingcrossmodal}, grounding supervision \citep{glip}, entity-level contrastive objectives \citep{dac}, structured concept training \citep{svlc}, and modular encoder patching with synthetic captions \citep{clove2024}. Training-free approaches decompose images and captions into constituent parts for local alignment \citep{comclip, abeclip}. Despite these efforts, gains remain modest and inconsistent, and recent work shows that improvements from hard-negative fine-tuning are significantly overstated as existing benchmarks fail to probe model invariance to hard positives \citep{kamath2024hardpositivetruthvisionlanguage}. Hence, it is crucial to understand where improvement actually occurs, whether models become better at binding specific primitive types or at reasoning over them jointly. COMPASS provides the controlled structure needed to answer these questions, enabling precise attribution of compositional gains and failures to specific primitive types and complexity levels.

\begin{table}[t!]
\setlength{\tabcolsep}{6pt}
\renewcommand{\arraystretch}{1.0}
\begin{center}
\resizebox{\columnwidth}{!}
{
\begin{tabular}{llclclccc}
\multicolumn{1}{c}{\textbf{}}                 &  & \textbf{}             &  & \multicolumn{5}{c}{\textbf{Evaluation Set}}                                     \\ \cmidrule{5-9} 
\multicolumn{1}{c}{}                          &  & \textbf{Ground Truth} &  & \textbf{Compositional Integration} &  & \multicolumn{3}{c}{\textbf{Skill Load}} \\ \cmidrule{3-3} \cmidrule{5-5} \cmidrule{7-9} 
\multicolumn{1}{c}{\textbf{Structural Level}} &  & Composed              &  & Composed/Decomposed                &  & Object     & Attribute    & Relation    \\ \midrule
L3 (OAR)                                      &  & 47K                   &  & 24K                                &  & 45K        & 26K          & 34K         \\
L2 (OA)                                       &  & 46K                   &  & 30K                                &  & 39K        & 37K          & -           \\
L2 (OR)                                       &  & 45K                   &  & 33K                                &  & 43K        & -            & 34K         \\ \bottomrule
\end{tabular}
}
\end{center}
\caption{\label{tab:stats}\textbf{\name{} Statistics.} We summarize the total size of our compositionality testbed across complexity levels with ground truth captions and evaluation set sizes for both compositional integration and skill load.}
\end{table}
\section{\name{}: A Testbed for Analysing Compositional Failures}
\label{sec:compass}


To diagnose the factors underlying compositional failure in VLMs as caption complexity increases, where complexity is defined as the total number of primitives (productivity in~\citealp{ma2023crepe}), we introduce \textbf{\name{}}, 
a controlled evaluation testbed. To enable this, we construct captions from scene graphs with explicit object, attribute, and relation structure, organizing them into hierarchical structural levels with systematic variation (Section~\ref{sec:gt_captions}). Models are evaluated using a retrieval-based protocol with hard negatives that isolate errors in recognizing specific primitive types (Section~\ref{sec:retrieval}). This controlled structure supports two targeted analyses: \textbf{compositional integration gap} (Section~\ref{sec:integration}), which isolates the cost of joint reasoning by comparing composed and decomposed captions under matched perturbations, and \textbf{skill load} (Section~\ref{sec:skill_load}), which measures how each skill degrades based on different primitive types.

\subsection{Structured Caption Construction}
\label{sec:gt_captions}

\begin{figure*}[t]
     \centering
     \includegraphics[width=\linewidth, trim={0cm 8.6cm 0cm 4cm},clip]{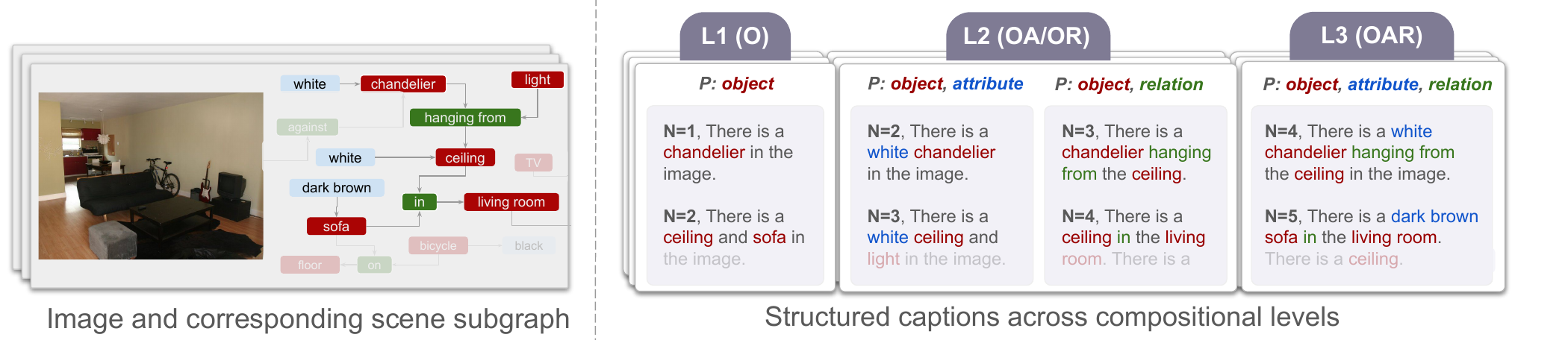}
     \caption{\textbf{Structured levels in \name{}.} With a subgraph of the scene graph of an input image, we construct captions organized into structured levels based on combinations of primitives (P): L1 (O: objects), L2 (OA/OR: objects with attributes or relations), and L3 (OAR: objects, attributes, and relations). This organization enables controlled evaluation of compositional reasoning by systematically varying object (O), attribute (A), and relation (R) components across levels.}
     \label{fig:data_structure}
 \end{figure*}

For controlled data construction, we use scene graphs from Visual Genome~\citep{visualgenome}, where each scene graph consists of objects (nodes), their attributes, and pairwise relations (edges) to generate captions.
For each image, we sample a subgraph $S$ using a random walk of up to ten steps (Fig.~\ref{fig:data_structure}), starting from a random object and traversing relation edges to expand the subgraph. This ensures a coherent subset of objects and relations with sufficient primitives to construct higher-complexity captions, while maintaining a controlled set of primitives that are largely reused across captions. Including an object automatically incorporates its associated attributes. The resulting subgraph serves as a basis for generating captions across different compositional structures and complexity levels, ensuring that the same set of primitives is reused across evaluations for a given image. COMPASS is built on \textit{5K} image -- scene graph pairs from Visual Genome. 

\textbf{Structural Levels and Complexity Control.}
To disentangle how the different primitive types contribute to compositional failure, we organize captions into hierarchical structural levels. This design allows us to isolate the impact of specific skills such as object recognition, attribute binding, and relation reasoning, which are often entangled in standard benchmarks. 
 We define captions in terms of primitives corresponding to \texttt{\obj{objects (O)}}, \texttt{\attr{attributes (A)}}, and \texttt{\rel{relations (R)}}. Structural levels are determined by the types of primitives present, as illustrated in Fig.~\ref{fig:data_structure}: L1 includes only objects, L2 includes objects with either attributes (OA) or relations (OR), and L3 includes objects, attributes, and relations jointly (OAR). For each structural level, we vary caption complexity by controlling the total number of primitives $N$. Specifically, for L1, $N \in [1, 10]$; for L2 (OA), $N \in [2, 12]$; for L2 (OR), $N \in [3, 12]$; and for L3, $N \in [4, 12]$. The lower limit is defined by the minimum number of primitives of different types required to create a valid caption. As we examine the composition of different primitives, we focus on L2 and L3 in this work. 
 
\textbf{Generating Natural Language Captions.}
Given this structured representation, we generate natural language captions that realize these primitives across different levels and complexities.
For a given complexity $N$ and structural level defined by primitive types $t \in \{O, A, R\}$, we traverse $S$ starting from a random node and incrementally add primitives until the caption $c$ contains $N$ elements, i.e., $P(c) = \{p_1, \dots, p_N\}$. We denote the count of each primitive type by $n_t(c)$. This construction allows controlled variation in both the composition and the number of components within each caption. For the final step of synthesis, we convert the resulting primitive sets into natural language captions using GPT 4o-mini ~\citep{gpt4o-mini} with few-shot examples (see Appendix). For example, as shown in Fig.~\ref{fig:data_structure}, ``\texttt{There is a \attr{dark brown} \obj{sofa} \rel{in} the \obj{living room}. There is a \obj{ceiling}.}'' is an L3 caption with complexity $N=5$, consisting of $n_o(c)=3$ (sofa, living room, ceiling), $n_r(c)=1$ (in), and $n_a(c)=1$ (dark brown). This procedure yields a total of \textbf{1.38M composed ground-truth captions} across structural levels (Table~\ref{tab:stats}). This structured caption space provides a controlled testbed for evaluating compositional behavior, using shared ground-truth captions across evaluations and retrieval-based evaluation using hard negatives (Section~\ref{sec:retrieval}) across models.

\subsection{Retrieval-based Evaluation using Hard Negatives}
\label{sec:retrieval}

Compositional understanding was evaluated using an image-to-text retrieval task, where a model must identify the ground-truth caption $c$ for an image among a set of candidate captions consisting of $c$ and hard negatives~\citep{hsieh2023sugarcrepe, ma2023crepe, aro}. Existing benchmarks have been shown to contain distributional artifacts between positive and hard-negative captions, rendering them exploitable by text-only models~\citep{hsieh2023sugarcrepe}. Addressing this fully requires a three-stage pipeline of LLM-based generation, human validation, and adversarial refinement~\citep{hsieh2023sugarcrepe}, which is not feasible at the scale required for COMPASS. We instead adopt an LLM-based substitution approach and conduct a perplexity audit to verify linguistic indistinguishability; full details are provided in the Appendix.

\textbf{Hard Negatives Generation.}
Given a caption $c$ with primitive set $P(c)$, we construct hard negatives by replacing a single primitive $p_i \in P(c)$ with another primitive of the same type $t$, while keeping all other primitives fixed as introduced in \cite{shekhar2017foil}. This results in captions that differ minimally and isolate errors in recognizing specific components. 
To generate replacements, we use GPT-4o mini with few-shot prompting to propose candidate primitives of the same semantic category (e.g., replacing ``\texttt{sofa}'' with ``\texttt{chair}''). We randomly sample a replacement that does not already appear in the image's entire scene graph.
We enforce additional constraints to ensure that negatives remain fluent and semantically plausible. After substitution, we adjust the caption for grammatical correctness and filter out overly similar candidates using Sentence Transformers~\citep{sentencetransformer}. This ensures that negatives are both natural and sufficiently distinct from the ground truth. (Refer to Appendix for more details)
\textbf{Example.}
For the caption ``\texttt{There is a \attr{dark brown} \underline{\obj{sofa}} \rel{in} the \obj{living room}. There is a \obj{ceiling}.}'', a corresponding hard negative is ``\texttt{There is a \attr{dark brown} \underline{\obj{chair}} \rel{in} the \obj{living room}. There is a \obj{ceiling}.}'', as shown in Figure~\ref{fig:CompDecomp}. 

This retrieval formulation provides the foundation for the evaluation settings introduced in Sections~\ref{sec:integration} (compositional integration) and ~\ref{sec:skill_load} (skill load), with specific hard negatives detailed in each section.

\section{Experimental Setup}

\label{sec:models}

To analyze compositional failure across different model families, we evaluate a diverse set of 
VLMs spanning standard architectures, recent variants, and models explicitly designed for compositional reasoning. Our selection includes \textbf{OpenCLIP}~\citep{openclip} (ViT-g/14, LAION-2B) as a standard contrastive baseline, along with recent variants such as \textbf{SigLIP v2}~\citep{tschannen2025siglip2multilingualvisionlanguage} and \textbf{PE-CLIP}~\citep{bolya2025PerceptionEncoder} (Perception Encoder). We further include compositional models: \textbf{NegCLIP}~\citep{aro}, which targets compositional generalization through hard-negative training, and \textbf{CE-CLIP}~\citep{zhang2024contrastingintramodalrankingcrossmodal}, which enhances compositional understanding by contrasting intra-modal and cross-modal hard negatives. We also evaluate \textbf{BLIP-L}~\citep{blip}, which bootstraps vision-language pretraining through a captioning and filtering pipeline and \textbf{Qwen3-VL-Embedding-8B}~\citep{li2026qwen3vlembeddingqwen3vlrerankerunifiedframework}, a high-capacity multimodal embedding model, to examine whether skill load effects extend beyond contrastive architectures.

All models are evaluated under a unified retrieval protocol (Section~\ref{sec:retrieval}). Given an image $I$ and a set of candidate captions, we compute image and text embeddings and rank candidates using similarity scores $s(I, c)$. This ensures a consistent evaluation setting across all architectures, enabling direct comparison of compositional integration and skill load across models. All models are evaluated on a single NVIDIA A100 GPU.




\section{Compositional Integration Gap: Composed vs. Decomposed}
\label{sec:integration}

To isolate the difficulty of joint reasoning, we compare model performance on a single composed caption against its performance on a set of independent, decomposed primitive captions. If models were perfectly compositional, performance should be invariant to whether primitives are presented jointly or in isolation. Any observed drop in the composed setting directly quantifies the cost of joint reasoning, which we define as the \textbf{compositional integration gap}.

\subsection{Data Setup}

\begin{figure*}[t]
     \centering
     \includegraphics[width=\linewidth, trim={0cm 3.5cm 0cm 3.3cm},clip]{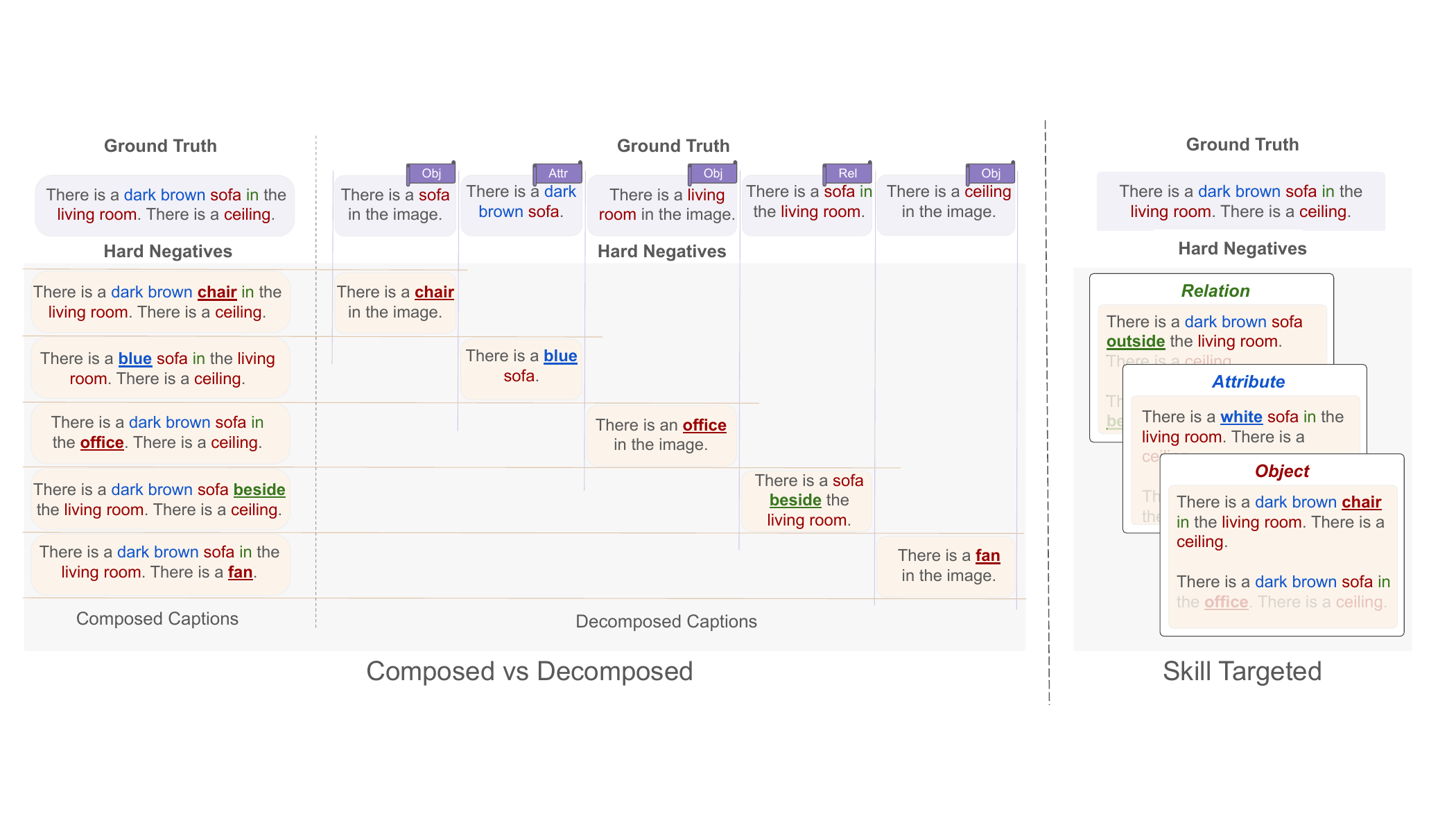}
     \caption{We construct two complementary negative structures to analyze compositional integration and skill load. \textbf{Left: Composed vs decomposed captions.} A composed caption with complexity $N$ is paired with $N$ hard negatives and decomposed into $N$ primitive captions. To enable controlled comparison between joint and independent reasoning, we apply the same perturbation to create negatives for both composed and decomposed. \textbf{Right: Skill-targeted negatives.} To analyze skill load, we isolate one skill at a time and create four hard negatives by modifying that primitive type. The figure shows an example from L3 with complexity $N=5$; this construction generalizes across all structural levels and complexities.}
     \label{fig:CompDecomp}
 \end{figure*}

\textbf{Decomposed Primitive Captions.} 
For every ground-truth caption $c$, we generate a set of $N$ decomposed captions $\{d_1, \dots, d_N\}$, where each $d_i$ targets exactly one primitive from the original scene graph. As illustrated in Fig.~\ref{fig:CompDecomp}, we define the structure of $d_i$ based on its primitive type: objects use a simple existential caption as they can be identified by themselves (e.g., ``\texttt{There is a \underline{\obj{sofa}} in the image.}''); attributes use an object-attribute pair for grounding (e.g., ``\texttt{There is a \underline{\attr{dark brown}} \obj{sofa}.}''); and relations use a triplet consisting of the relation and its two bridging objects (e.g., ``\texttt{The \obj{sofa} is \underline{\rel{in}} the \obj{living room}.}'').

\textbf{Matched Perturbations for Task Equivalence.} 
To ensure a rigorous comparison, we maintain a strict one-to-one correspondence between the negatives used in the composed and decomposed settings. For a composed caption $c$ with complexity $N$, we generate $N$ hard negatives $\{\tilde{c}_1, \dots, \tilde{c}_N\}$, where each negative $\tilde{c}_i$ is formed by perturbing exactly one primitive $p_i \in P(c)$. Each decomposed caption $d_i$ is then paired with a negative $\tilde{d}_i$ created using the exact same perturbation applied to its corresponding composed negative $\tilde{c}_i$. 
\textbf{Example.}
As shown in Figure~\ref{fig:CompDecomp}, if a composed caption changes the object `\texttt{\obj{sofa}}' to `\texttt{\obj{chair}}' to create a hard negative, the corresponding decomposed caption, ``\texttt{There is a \underline{\obj{sofa}} in the image.}'' is paired with the exact same semantic perturbation ``\texttt{There is a \underline{\obj{chair}} in the image.}'' 

While a composed negative perturbs a primitive within its full context, the decomposed instance isolates that same primitive by removing the surrounding context from the prompt. 
To correctly retrieve the ground-truth composed caption, a model must implicitly resolve the correctness of every constituent primitive. Our decomposed setting makes this requirement explicit: success requires the model to retrieve every decomposed primitive against its respective negative independently. We construct \textbf{87K} caption pairs across structural levels to evaluate compositional integration (Table~\ref{tab:stats}).

\subsection{Metrics for Compositional Integration.} \label{sec:CI_metrics}
We formally define the \textbf{Compositional Integration Gap} ($\Delta$) as the difference between the model's ability to retrieve the correct caption in the decomposed versus composed settings. For a composed caption $c$, Recall@1 is defined as:
\begin{equation}
\label{eq:recall_comp}
\text{R@1}_{\text{comp}}(c) = \mathbf{1}\!\left[s(I,c) > s(I,\tilde{c}_i), \ \forall i \in \{1, \dots, N\} \right].
\end{equation}
For the decomposed setting, we aggregate performance using a joint success criterion, requiring the model to correctly retrieve every single primitive in the set:
\begin{equation}
\label{eq:recall_decomp}
\text{R@1}_{\text{decomp}}(c) = \prod_{i=1}^{N} \mathbf{1}\!\left[s(I,d_i) > s(I,\tilde{d}_i)\right].
\end{equation}
The compositional integration gap $\Delta$ is then defined as the difference between the decomposed (Eq.~\ref{eq:recall_decomp}) and composed (Eq.~\ref{eq:recall_comp}) recall scores:
\begin{equation}
\label{eq:delta}
\Delta(c) = \text{R@1}_{\text{decomp}}(c) - \text{R@1}_{\text{comp}}(c).
\end{equation}
This formulation allows us to determine if model failure stems from the inherent complexity of joint reasoning or from the failure to recognize individual components.

\subsection{How Much Does Joint Reasoning Cost?}
\label{sec:CI_results}

Table~\ref{tab:integration} reports the mean and standard deviation of $\Delta$ aggregated across all complexities, with detailed results in the Appendix. Across models and structural levels, $\Delta$ is predominantly positive, confirming that joint reasoning introduces a measurable cost over independent primitive recognition.

Among standard and recent contrastive models, PE-CLIP shows the smallest integration gaps across all structural levels, suggesting greater robustness to compositional binding, while SigLIPv2 exhibits the largest gaps among this group, indicating strong individual primitive recognition that does not transfer to joint reasoning. Among non-contrastive models, BLIP-L shows moderate gaps consistent with the contrastive baseline, while Qwen3 exhibits the largest OAR gap overall, suggesting that even high-capacity multimodal embedding models struggle with compositional integration despite their representational power.

NegCLIP and CE-CLIP are notable exceptions, both achieving negative $\Delta$ values, indicating that composed performance exceeds decomposed performance. NegCLIP shows this behavior only on OA, while CE-CLIP shows it consistently across all structural levels. This suggests that explicit compositional training enables these models to leverage joint context across primitives more effectively, such that composed captions are actually easier to retrieve than their decomposed counterparts. This qualitatively different behavior suggests that compositional training objectives can improve joint reasoning rather than merely shifting distributional preferences.
\begin{table}[t!]
\centering
\resizebox{\columnwidth}{!}{
\setlength{\tabcolsep}{6pt}
\renewcommand{\arraystretch}{1.1}

\begin{tabular}{llccccccl}
\toprule
\textbf{Type} &  & \textbf{OpenCLIP} & \textbf{SigLIPv2} & \textbf{PE-CLIP} & \textbf{NegCLIP} & \textbf{CE-CLIP} & \textbf{BLIP-L} & \textbf{Qwen3}                    \\ \midrule
OAR           &  & 0.83$\pm$0.70     & 2.61$\pm$1.82    & 1.97$\pm$0.59    & 3.75$\pm$0.64    & \sr{-5.66$\pm$3.45}  & 1.38$\pm$1.03   & \multicolumn{1}{l}{5.34$\pm$4.42} \\
OA            &  & 2.39$\pm$1.23     & 12.65$\pm$2.53    & 0.94$\pm$1.96    & \sr{-2.70$\pm$1.83}   & \sr{-20.27$\pm$5.17}  & 1.19$\pm$2.17   & \multicolumn{1}{l}{11.94$\pm$4.81}              \\
OR            &  & 17.56$\pm$1.98     & 8.64$\pm$5.04     & 2.69$\pm$0.70    & 5.54$\pm$1.44   & \sr{-4.96$\pm$3.93}  & 2.30$\pm$2.93   &          7.31$\pm$5.16                        \\ \bottomrule \\
\end{tabular}
}
\caption{\label{tab:integration}\textbf{Compositional Integration (Aggregated).}  We report the mean $\pm$ standard deviation of the difference between decomposed and composed accuracy ($\Delta$) aggregated across all complexities. Positive values indicate better independent than joint performance, highlighting compositional binding difficulty. Shaded cells indicate models for which composed captions are easier to retrieve than their decomposed counterparts.}
\end{table}

\begin{wrapfigure}{r}{0.42\textwidth}
    \centering
    \includegraphics[width=0.42\textwidth, trim={0cm 0cm 0cm 0cm}, clip]{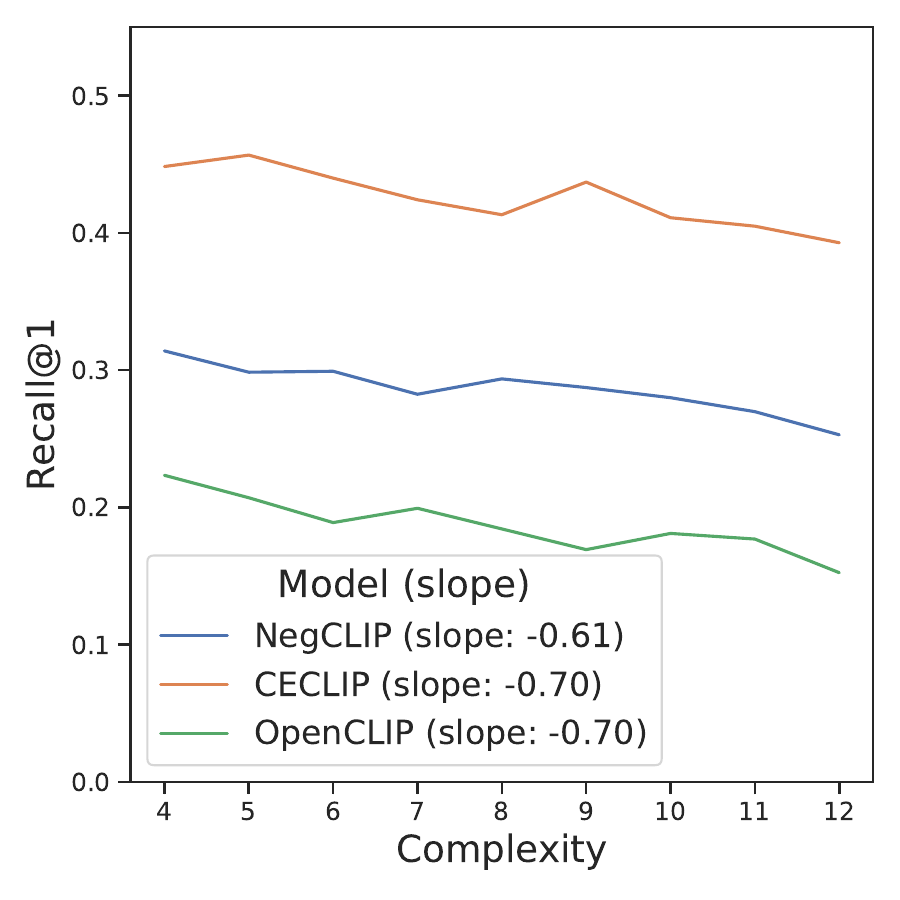}
    \caption{Performance on CREPE}
    \label{fig:crepe_productivity}
\end{wrapfigure}
\textbf{Discussion.} The compositional integration gap provides a useful upper bound on model capability, showing how well models perform when each primitive is evaluated in isolation. While compositional training objectives such as NegCLIP and CE-CLIP can reverse the integration gap, making composed retrieval easier than decomposed retrieval, both models, like CLIP, still exhibit monotonic performance degradation with increasing caption complexity on CREPE's productivity benchmark (Figure~\ref{fig:crepe_productivity}, refer to the appendix for more details). This raises a deeper question: \textit{what drives performance degradation as caption complexity grows, beyond joint reasoning alone?} Prior work has treated complexity as a single scalar, without distinguishing the contributions of individual primitive types. To answer this, we move beyond this aggregate view and analyze how model performance varies with the counts of each primitive type independently, discussed next in Section~\ref{sec:skill_load}.

\section{Skill Load: Skill Targeted Perturbations}
\label{sec:skill_load}
\begin{table}[t!]
\setlength{\tabcolsep}{7pt}
\renewcommand{\arraystretch}{1.0}
\begin{center}
\resizebox{\columnwidth}{!}
{
\begin{tabular}{llllccccccccccc}
\toprule[2pt]
\textbf{Type} &  & \textbf{Model} &  & \multicolumn{3}{c}{\textbf{Object Negatives}}         & \multicolumn{1}{l}{} & \multicolumn{3}{c}{\textbf{Attribute Negatives}}      & \multicolumn{1}{c}{} & \multicolumn{3}{c}{\textbf{Relation Negatives}}       \\ \cmidrule{5-7} \cmidrule{9-11} \cmidrule{13-15} 
              &  &                &  & $\beta_O$ & \multicolumn{1}{c}{$\beta_A$} & $\beta_R$ & \multicolumn{1}{c}{} & $\beta_O$ & \multicolumn{1}{c}{$\beta_A$} & $\beta_R$ & \multicolumn{1}{c}{} & $\beta_O$ & \multicolumn{1}{c}{$\beta_A$} & $\beta_R$ \\ \midrule
\textbf{OAR}  &  & OpenCLIP           &  & \sr{-2.50**}   & +1.21**                       & +2.32**   &                      & +1.76**   & \sr{-3.32**}                       & -1.84     &                      & +0.60     & +1.17**                       & \sr{-2.86**}   \\
              &  & SigLIP 2       &  & \sr{-2.25**}   & +0.65*                        & +1.82     &                      & +2.60**   & \sr{-2.85**}                       & -3.07     &                      & +0.70     & -0.77**                       & \sr{-1.65*}    \\
              &  & PE-CLIP        &  & \sr{-2.45**}   & +0.96**                       & +1.39     &                      & +1.07     & \sr{-4.13**}                       & -0.50     &                      & +0.40     & -0.09                         & \sr{-1.91*}    \\
              &  & NegCLIP        &  & \sr{-1.762**}  & +0.69                         & +0.81**   &                      & +1.51**   & \sr{-3.38**}                       & -2.45     &                      & +2.31**   & +0.13                         & \sr{-7.13**}   \\
              &  & CE-CLIP        &  & \sr{-1.77**}   & +0.68**                       & +2.23*    &                      & +0.20**   & \sr{-2.57**}                       & -2.81     &                      & -1.08     & +2.31**                       & \sr{-4.89**}   \\
              &  & BLIP-L         &  & \sr{-3.55**}   & +1.08**                       & +3.72     &                      & +0.50     & \sr{-2.92**}                       & +0.5      &                      & +0.2      & -1.32                         & \sr{-2.95**}   \\
              &  & Qwen3   &  & \sr{-1.87**}   & +1.25**                       & +0.87     &                      & +1.95**   & -3.17**                      & \sr{-3.71**}   &                      & +0.64     & -0.51                         & \sr{-2.66**}   \\ \midrule
\textbf{OA}   &  & OpenCLIP           &  & \sr{-1.46**}   & -0.82*                        & -         &                      & +0.73**   & \sr{-3.76**}                       & -         &                      & -         & -                             & -         \\
              &  & SigLIP 2       &  & \sr{-2.27**}   & -1.35**                       &           &                      & -0.01     & \sr{-3.21**}                       &           &                      & -         & -                             & -         \\
              &  & PE-CLIP        &  & \sr{-1.98**}   & -0.52                         & -         &                      & +0.10     & \sr{-4.00**}                       & -         &                      & -         & -                             & -         \\
              &  & NegCLIP        &  & \sr{-2.12**}   & -1.08**                       & -         &                      & -0.08     & \sr{-3.72**}                       & -         &                      & -         & -                             & -         \\
              &  & CE-CLIP        &  & -2.62     & \sr{-0.73*}                        & -         &                      & +1.21**   & \sr{-2.37**}                       & -         &                      & -         & -                             & -         \\
              &  & BLIP-L         &  & \sr{-0.50**}   & +0.03                         & -         &                      & +0.50**   & \sr{-2.53**}                       & -         &                      & -         & -                             & -         \\
              &  & Qwen3   &  & \sr{-1.60**}   & +0.07                         & -         &                      & -0.13     & \sr{-3.31**}                       & -         &                      & -         & -                             & -         \\ \midrule
\textbf{OR}   &  & OpenCLIP           &  & \sr{-1.45**}   & -                             & +1.89*    &                      & -         & -                             & -         &                      & +1.05**   & -                             & \sr{-2.42**}   \\
              &  & SigLIP 2       &  & \sr{-1.89**}   & -                             & +2.22**   &                      & -         & -                             & -         &                      & +0.45     & -                             & -1.28     \\
              &  & PE-CLIP        &  & \sr{-2.03**}   & -                             & +1.49     &                      & -         & -                             & -         &                      & +0.50     & -                             & \sr{-1.69*}    \\
              &  & NegCLIP        &  & \sr{-0.95*}    & -                             & +0.65     &                      & -         & -                             & -         &                      & +2.29**   & -                             & \sr{-6.81**}   \\
              &  & CE-CLIP        &  & \sr{-2.15**}   & -                             & +3.98**   &                      & -         & -                             & -         &                      & +1.01     & -                             & \sr{-4.68**}   \\
              &  & BLIP-L         &  & \sr{-2.53**}   & -                             & +3.46**   &                      & -         & -                             & -         &                      & \sr{-1.30**}   & -                             & +0.80     \\
              &  & Qwen3   &  & \sr{-1.27**}   & -                             & +1.25     &                      & -         & -                             & -         &                      & +0.47     & -                             & -1.90     \\ \bottomrule
\end{tabular}
}
\end{center}
\caption{\textbf{Skill Load. }Values represent change in R@1 (percentage points) per unit increase in primitive count (Eq.~\ref{eq:skill_load}). Negative values indicate degradation, positive values indicate improvement. Shaded cells indicate the dominant source of degradation within each setting. * and ** denote statistical significance at p < 0.005 and p < 0.001, respectively. Object, attribute, and relation self-load are consistently negative across models while cross-load effects are predominantly positive. \label{tab:skill_load}}
\end{table}

Building on the discussion above, we turn to a more fine-grained analysis of what drives performance degradation in composed captions. Specifically, we examine how performance on each skill---object, attribute, and relation---varies with the number of primitives of each type—what we term \textbf{\textit{skill load}}. We distinguish between \textbf{\textit{self-load}}, the degradation induced by increasing the count of the same primitive type being tested, and \textbf{\textit{cross-load}}, the degradation induced by increasing the count of other primitive types.

\subsection{Data Setup}
To isolate effects on individual skills, we construct skill-targeted hard negatives. For a given primitive type $t \in \{O, A, R\}$, we replace primitives of type $t$ only, keeping all other primitives fixed, allowing us to measure a model's ability to detect a specific primitive type independently of perturbations to others. We generate $K=4$ hard negatives per caption, as \citet{udandarao2025goodcrepeneedsjust} shows that single-negative retrieval introduces evaluation instability and underestimates model failures. To maintain consistency across skills, we enforce that no two negatives for the same caption are identical; where a caption contains fewer than $K$ distinct primitives of type $t$, the same primitive may be targeted across multiple negatives subject to this constraint. This construction generalizes across all structural levels and complexities. 
\textbf{Example.} As shown in Figure~\ref{fig:CompDecomp}, for the caption ``\texttt{There is a \attr{dark brown} \obj{sofa} \rel{in} the \obj{living room}. There is a \obj{ceiling}.}'', an object negative would be ``\texttt{There is a \attr{dark brown} \underline{\obj{chair}} \rel{in} the \obj{living room}. There is a \obj{ceiling}.}'', an attribute negative would be ``\texttt{There is a \underline{\attr{white}} \obj{sofa} \rel{in} the \obj{living room}. There is a \obj{ceiling}.}'' This construction yields \textbf{143K} ground-truth caption pairs for object evaluation, \textbf{63K} for attribute, and \textbf{68K} for relation, across structural levels (Table~\ref{tab:stats}).

\subsection{Metrics}
\label{sec:skill_load-metrics}

We measure skill load by modeling how R@1, computed over skill-targeted negatives for primitive type $t$, varies with the counts of each primitive type. Formally:
\begin{equation}
\text{R@1}_t = \beta_O \, n_O(c) + \beta_A \, n_A(c) + \beta_R \, n_R(c) + \alpha
\label{eq:skill_load}
\end{equation}
where $n_t(c)$ denotes the number of primitives of type $t$ in caption $c$, $\beta_t$ denotes the change in performance per unit increase in $n_t(c)$ while controlling for other types, and $\alpha$ denotes the intercept term. A negative coefficient $\beta_t$ indicates that increasing the count of primitive type $t$ degrades performance on skill $t$. Comparing $\beta_t$ across primitive types allows us to distinguish self-load effects from cross-load effects for each skill. We estimate Equation~\ref{eq:skill_load} using OLS regression with standard errors clustered at the image level to account for within-image correlation. Statistical significance is assessed using two-sided t-tests at $p<0.005$ (denoted *) and $p<0.001$ (denoted **).

\subsection{What Drives Skill-Level Degradation?}
Table~\ref{tab:skill_load} reports the estimated coefficients from Eq.~\ref{eq:skill_load} across models and structural levels, with shaded cells indicating the dominant source of degradation with high significance within each setting. Visualization of the underlying degradation patterns 
across all models is provided in Appendix.

\textbf{Self-load dominates cross-load.} Across all models and structural levels, the largest and most consistent negative coefficients appear along the diagonal, meaning each skill degrades primarily with the count of its own primitive type. Cross-load effects are predominantly positive, driven primarily by mutual grounding between objects and attributes, as well as between objects and relations, suggesting that co-occurring primitives of different types provide useful context rather than competing for representational capacity. This pattern is consistent across model families, including standard contrastive models (OpenCLIP, SigLIPv2, PE-CLIP), models explicitly trained for compositional reasoning (NegCLIP, CE-CLIP), and non-contrastive models (BLIP-L, Qwen3), indicating that self-load degradation is a fundamental property of current vision-language architectures rather than an artifact of any particular training objective. Among the three skills, attribute self-load is the strongest effect, consistently significant at $p<0.001$ across all models and structural levels, indicating that attribute binding is particularly sensitive to load. Relation self-load is negative and significant across most models but smaller in magnitude, with positive cross-load from object count, suggesting that objects provide a grounding context that aids relation detection.

\textbf{Discussion.}
These findings show that compositional degradation is not uniform across skills and cannot be reduced to a single joint reasoning bottleneck: primitive types contribute unequally to performance degradation, and treating caption complexity as a single scalar obscures these distinctions. Performance degradation is largely decomposable into skill-specific self-load effects, with cross-load effects providing modest grounding benefits rather than additional interference. Notably, NegCLIP and CE-CLIP, which reverse the compositional integration gap (Section~\ref{sec:CI_results}), still exhibit consistent self-load degradation across all structural levels, suggesting that compositional training objectives address joint reasoning difficulty but leave the underlying load sensitivity of individual skills unresolved. 

\vspace{-2pt}
\section{Conclusion}
\label{sec:conclusion}

Despite extensive efforts to benchmark and improve compositional understanding in vision-language models, the sources of performance degradation remain poorly understood. Through \name{}, a controlled evaluation framework that constructs captions from scene graphs with explicit object, attribute, and relation structure, we show that the observed degradation with increasing caption complexity reflects multiple separable factors. Compositional integration introduces a measurable cost, yet accounts for only part of the observed degradation; even models that handle joint reasoning well continue to suffer as caption complexity grows. Skill load analysis reveals a more fundamental pattern: each skill degrades primarily under the weight of its own primitive count, with cross-load effects providing grounding benefits rather than additional interference, and this holds consistently across standard, compositional, and non-contrastive model families. These findings reframe compositional failure as a multi-factorial problem in which joint reasoning accounts for only part of the observed degradation. Progress on compositional robustness may therefore require skill-specific interventions that ensure individual skills remain robust as the primitive load increases, rather than focusing solely on joint reasoning mechanisms. \name{} provides a controlled structure for diagnosing these factors independently and tracking progress along each dimension.

\textbf{Limitations and Future Work.} Several limitations point to directions for future work. COMPASS is constructed from Visual Genome scene graphs using synthetically generated captions, following a long line of work that uses controlled synthetic data for compositional evaluation~\citep{ma2023crepe, johnson2016clevrdiagnosticdatasetcompositional}, which may not fully reflect the distribution of naturally occurring descriptive language, and inherits known annotation biases toward certain object and relation types; future work could extend COMPASS to naturally occurring captions and more diverse scene graph sources. Furthermore, because objects provide the necessary grounding context for both attributes and relations, our framework evaluates attribute and relation skills in the presence of objects rather than in complete isolation, which is an inherent constraint of scene-graph-based evaluation. Developing evaluation protocols that further disentangle these dependencies remains an open challenge. Finally, our retrieval-based evaluation protocol, which is standard in prior compositionality benchmarks~\citep{hsieh2023sugarcrepe, ma2023crepe, aro}, isolates discrimination ability but does not extend to generative VLM settings, where failure modes may differ; adapting COMPASS's controlled-complexity structure to generative evaluation is a natural direction for future work.
Beyond these methodological limitations, our findings also raise deeper questions about the origin of self-load degradation. The pattern in which each skill degrades specifically under the weight of its own primitive count, while cross-load effects are positive, is suggestive of a representational account: models may have limited capacity to maintain robust encodings of multiple instances of the same primitive type, with degradation reflecting competition within rather than across primitive types. Whether this reflects encoding failures, cross-modal alignment failures, or both remains an open question for future work.


\begin{ack}

\end{ack}

\bibliography{neurips}
\bibliographystyle{neurips}


\appendix
\section{Details About Hard Negative Generation}
\label{app:prompts}

\begin{figure*}[t]
     \centering
     \includegraphics[width=\linewidth, trim={1.2cm 0cm 2cm 0cm},clip]{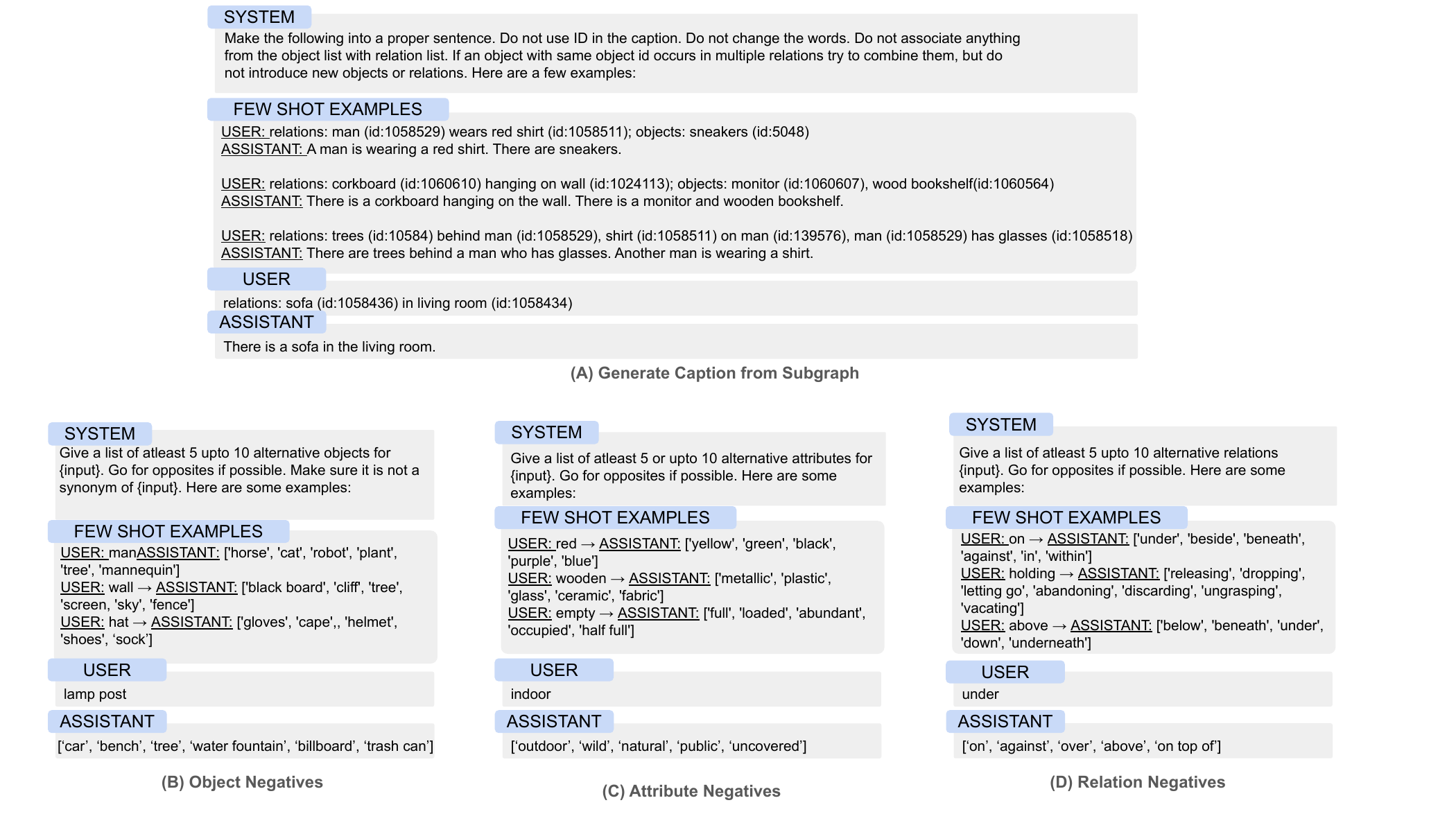}
     \caption{GPT-4o mini prompts used for caption and negative generation. (A) Few-shot prompt for generating natural language captions from scene subgraphs, instructing the model to combine primitives into fluent sentences without introducing new objects or relations. (B-D) Few-shot prompts for generating hard negative candidates for object (B), attribute (C), and relation (D) primitives respectively, eliciting semantically plausible alternatives that are neither synonyms nor identical to the original primitive.}
     \label{fig:gpt_queries}
 \end{figure*}

Figure~\ref{fig:gpt_queries} shows the GPT-4o mini prompts used for caption and negative generation, as described in Sections~\ref{sec:gt_captions} and~\ref{sec:retrieval}, respectively. For caption generation (A), we use a few-shot prompt that instructs the model to convert scene graph primitives into fluent natural language sentences, without introducing new objects or relations beyond those specified. For hard negative generation (B-D), we use separate few-shot prompts for each primitive type. Each prompt instructs the model to generate semantically plausible alternatives that are neither synonyms nor identical to the original primitive, with a preference for opposites where possible.

\textbf{Candidate caching.} To avoid redundant API calls, generated candidates are stored in a per-primitive-type dictionary mapping each primitive to its list of alternatives. When a negative is needed for a given primitive, we first check the dictionary; if no valid candidate exists, we query GPT-4o mini and append the results to the dictionary. This allows candidates to be reused across captions that share the same primitives.

\textbf{Filtering.} To construct a hard negative, we replace the target primitive with a candidate alternative and adjust the caption for grammatical correctness using Language Tool\footnote{\url{https://pypi.org/project/language-tool-python/}} before any filtering is applied. The resulting caption is then evaluated using Sentence Transformers~\citep{sentencetransformer} to ensure semantic distinctiveness from the original caption. A candidate is discarded if its similarity score to the original caption exceeds a threshold: we use $0.9$ for captions with complexity $N \geq 6$ and $0.95$ for $N < 6$. At lower complexities, a single word change has a larger impact on overall caption similarity, so we apply a stricter threshold to ensure candidates are sufficiently distinct. At higher complexities, the same substitution contributes less to the overall similarity score, requiring a more relaxed threshold to avoid discarding valid candidates. This filtering is applied consistently across composed, decomposed, and skill-targeted negatives.

\textbf{Primitive selection for skill-targeted negatives.} For skill-targeted negatives, primitives are selected using a weighted sampling scheme. Initially all primitives of the target type are assigned equal weights; once a primitive is selected, its weight is halved to encourage diversity across the $K=4$ negatives per caption. We make up to 10 attempts to generate 4 valid negatives per caption; captions for which 4 valid negatives cannot be obtained are discarded from evaluation.

\textbf{Composed and decomposed negatives.} For composed and decomposed settings, each negative must target exactly one primitive. Once a valid negative is obtained for a given primitive, that primitive is not used again for the same caption. We make up to 3 attempts to obtain a valid negative per primitive; if a valid negative cannot be found for all primitives in a caption, that instance is discarded from evaluation.

\section{Perplexity Analysis}
\label{app:perplexity}

\subsection{Overview}

To validate that COMPASS hard negatives cannot be distinguished from ground-truth captions using linguistic statistics alone, we conduct a systematic perplexity audit. Motivated by SugarCREPE~\cite{hsieh2023sugarcrepe}, which identifies perplexity-based exploitability as a critical flaw in substitution-based benchmarks, we measure the distributional difference between ground-truth and negative caption perplexities across all primitive types and structural levels. We compute token-level perplexity under GPT-2~\cite{gpt2} for all ground-truth captions and their corresponding hard negatives. A benchmark where negatives are systematically less fluent than ground-truth captions is exploitable by text-only models, as a model can identify the ground-truth caption by selecting the lowest-perplexity candidate without any visual input.

\subsection{Evaluating on the Hardest Negative}

For each ground-truth caption with $K=4$ skill-targeted hard negatives, we identify the \emph{hardest negative} — the candidate whose perplexity is closest to the ground-truth caption:

\begin{equation}
    \hat{n} = \arg\min_{n \in \mathcal{N}} \left| \text{PPL}(n) - 
    \text{PPL}(c) \right|
\end{equation}

This choice is motivated by the structure of our retrieval task. In $K=4$ evaluation, a model achieves R@1$=1$ only if the ground-truth caption scores higher than \emph{all four} negatives. For a text-only model exploiting perplexity to succeed on a trial, it must identify the ground-truth as having lower perplexity than all four negatives — including the one most similar in perplexity to the ground truth. If even this hardest-to-exploit negative is linguistically indistinguishable from the ground truth, then the full $K=4$ task cannot be reliably solved via perplexity alone. Measuring artifact severity on the hardest negative therefore represents a \emph{worst-case bound} on exploitability: if the effect size is negligible on this subset, it is negligible across the entire evaluation.

\subsection{Effect Size Metric}

\begin{equation}
    r = 1 - \frac{2U}{n_1 n_2}
\end{equation}
where $U$ is the Mann-Whitney statistic and $n_1$, $n_2$ are the sample sizes of the two groups. The rank-biserial correlation measures the probability that a randomly drawn ground-truth perplexity exceeds a randomly drawn negative perplexity, expressed as a deviation from chance. We interpret effect sizes following standard conventions: $r < 0.1$ indicates a negligible difference, $0.1 \leq r < 0.3$ indicates a small difference, and $r \geq 0.3$ indicates a medium or large difference. We use this metric rather than a text-only retrieval baseline because it directly characterizes the distributional overlap between ground-truth and negative perplexities, without dependence on task format or chance level definitions that vary across evaluation settings.

\subsection{Results: Skill-Targeted Negatives}
\label{app:perplexity_skill}

Table~\ref{tab:perplexity_skill} reports effect sizes for skill-targeted 
negatives across all structural levels, measured on the hardest negative 
subset.

\begin{table}[h]
\centering
\begin{tabular}{llcc}
\toprule
Level & Negative Type & Effect size $r$ & Interpretation \\
\midrule
L3 (OAR) & Object    & 0.096 & Negligible \\
L3 (OAR) & Attribute & 0.072 & Negligible \\
L3 (OAR) & Relation  & 0.259 & Small      \\
L2 (OA)  & Object    & 0.062 & Negligible \\
L2 (OA)  & Attribute & 0.099 & Negligible \\
L2 (OR)  & Object    & 0.099 & Negligible \\
L2 (OR)  & Relation  & 0.267 & Small      \\
\bottomrule \\
\end{tabular}
\caption{\label{tab:perplexity_skill}Perplexity effect sizes (rank-biserial $r$) measuring the distributional difference between ground-truth and negative caption perplexities for skill-targeted negatives, evaluated on the hardest negative subset.}
\end{table}

Object and attribute negatives exhibit negligible effect sizes across all structural levels ($r = 0.062$--$0.099$), confirming that ground-truth and negative captions are linguistically indistinguishable under worst-case selection. Relation negatives show slightly larger but still small effect sizes ($r = 0.259$--$0.267$), consistent with stronger collocational constraints on spatial prepositions compared to content words. Across all primitive types and structural levels, effect sizes remain within the small range by standard conventions, validating the skill load results reported in Section~\ref{sec:skill_load}.

\subsection{Results: Compositional Integration Negatives}
\label{app:perplexity_integration}

Table~\ref{tab:perplexity_integration} reports effect sizes for composed captions across structural levels, measured on the hardest negative subset.

\begin{table}[h]
\centering
\begin{tabular}{lcc}
\toprule
Level & Effect size $r$ & Interpretation \\
\midrule
L3 (OAR) & 0.060 & Negligible \\
L2 (OA)  & 0.152 & Small      \\
L2 (OR)  & 0.071 & Negligible \\
\bottomrule \\
\end{tabular}
\caption{\label{tab:perplexity_integration}Perplexity effect sizes (rank-biserial $r$) measuring the distributional difference between ground-truth and negative caption perplexities for composed captions, evaluated on the hardest negative subset.}
\end{table}

Composed captions exhibit negligible to small perplexity artifacts across all structural levels ($r = 0.060$--$0.152$), confirming that the composed evaluation setting is not systematically exploitable via linguistic statistics. This validates the composed side of the compositional integration gap analysis reported in Section~\ref{sec:integration}.

\paragraph{Decomposed perplexity results.}
Decomposed captions are structurally short by design - each targets one primitive, yielding captions of typically 6--8 tokens (e.g., ``There is a sofa in the image.''). At this length, substituting a single word constitutes a proportionally large fraction of the caption's total token sequence, and perplexity is dominated by the corpus frequency of the substituted word rather than any meaningful fluency difference. These artifacts are a structural consequence of caption length rather than the negative generation pipeline, and decomposed perplexity results are therefore not informative for benchmark validity. The composed setting, which forms one side of the integration gap computation, is artifact-free across all structural levels ($r = 0.060$--$0.152$).

\section{Compositional Integration Gap Raw Results}
The compositional integration gap $\Delta$ measures the difference between decomposed and composed accuracy, quantifying the additional cost introduced by joint reasoning over multiple primitives. A positive $\Delta$ indicates that models perform better when primitives are evaluated independently than when they must be resolved jointly. We report per-complexity results across all structural levels here; aggregate results are discussed in Section~\ref{sec:CI_results}.

\begin{table}[h]
\centering
\scriptsize
\setlength{\tabcolsep}{2.2pt}
\resizebox{\columnwidth}{!}{
\begin{tabular}{@{}c|ccc|ccc|ccc|ccc|ccc|ccc|ccc@{}}
\toprule

 & \multicolumn{3}{c|}{\textbf{OpenCLIP}} & 
\multicolumn{3}{c|}{\textbf{SigLIP 2}} & 
\multicolumn{3}{c|}{\textbf{PE-CLIP}} & 
\multicolumn{3}{c|}{\textbf{NegCLIP}} & 
\multicolumn{3}{c|}{\textbf{CE-CLIP}} & 
\multicolumn{3}{c}{\textbf{BLIP-L}} & 
\multicolumn{3}{c}{\textbf{Qwen3}} \\

 \textbf{Complexity Level} & \textbf{C} & \textbf{D} & 
\textbf{$\Delta$} & \textbf{C} & \textbf{D} & 
\textbf{$\Delta$} & \textbf{C} & \textbf{D} & 
\textbf{$\Delta$} & \textbf{C} & \textbf{D} & 
\textbf{$\Delta$} & \textbf{C} & \textbf{D} & 
\textbf{$\Delta$} & \textbf{C} & \textbf{D} & 
\textbf{$\Delta$} & \textbf{C} & \textbf{D} & 
\textbf{$\Delta$} \\ \midrule

 2 & 55.69 & 55.26 & -0.42 & 58.70 & 65.66 & 6.96 & 65.48 & 62.20 & -3.28 & 62.08 & 56.61 & -5.47 & 62.90 & 56.14 & -6.75 & 63.51 & 64.37 & 0.85 & 50.38 & 63.87 & 13.49 \\

 3 & 43.42 & 47.55 & 4.13 & 41.33 & 56.19 & 14.86 & 54.79 & 54.23 & -0.56 & 50.40 & 46.69 & -3.71 & 57.98 & 42.51 & -15.46 & 47.94 & 53.14 & 5.19 & 34.7	& 53.43	& 18.72 \\

 4 & 35.13 & 37.01 & 1.87 & 34.04 & 47.78 & 13.75 & 45.61 & 44.12 & -1.49 & 39.57 & 35.47 & -4.11 & 51.76 & 31.58 & -20.17 & 39.86 & 44.89 & 5.02 & 26.43 & 43.29 &	16.86\\

 5 & 29.61 & 31.77 & 2.16 & 28.77 & 44.00 & 15.23 & 38.23 & 39.83 & 1.60 & 33.12 & 31.37 & -1.75 & 48.68 & 26.08 & -22.60 & 34.47 & 36.72 & 2.25 & 19.71 & 36.64 &	16.93 \\

 6 & 22.71 & 25.51 & 2.79 & 21.94 & 36.91 & 14.96 & 30.64 & 31.95 & 1.31 & 28.75 & 25.18 & -3.57 & 44.18 & 20.54 & -23.64 & 29.01 & 30.46 & 1.44 & 17.29 & 31.45 & 14.16\\

 7 & 18.82 & 21.00 & 2.17 & 18.61 & 32.32 & 13.71 & 26.32 & 27.29 & 0.97 & 23.25 & 19.93 & -3.32 & 42.01 & 16.65 & -25.35 & 25.17 & 26.17 & 1.00 & 14.6	& 26.97 & 12.37 \\

 8 & 15.33 & 19.00 & 3.67 & 16.49 & 30.27 & 13.78 & 22.45 & 25.28 & 2.83 & 19.28 & 15.91 & -3.37 & 36.45 & 13.11 & -23.33 & 22.45 & 21.83 & -0.61 & 12.3 & 23.58 & 11.27\\

 9 & 13.01 & 14.55 & 1.53 & 12.56 & 26.04 & 13.47 & 18.57 & 21.77 & 3.19 & 16.75 & 13.97 & -2.78 & 33.47 & 10.50 & -22.96 & 20.06 & 19.44 & -0.62 & 10.66 & 20.76 &	10.11\\

 10 & 10.64 & 13.72 & 3.07 & 11.39 & 22.03 & 10.64 & 16.40 & 18.69 & 2.29 & 11.39 & 11.43 & 0.04 & 29.37 & 8.05 & -21.32 & 16.44 & 16.27 & -0.17 & 10.17 & 18.57 &	8.40\\

 11 & 8.03 & 11.29 & 3.25 & 9.19 & 19.71 & 10.53 & 14.35 & 16.08 & 1.72 & 11.82 & 9.23 & -2.58 & 28.01 & 6.96 & -21.05 & 14.44 & 14.16 & -0.28 & 9.68 & 13.45 &	3.77\\

 12 & 7.46 & 9.59 & 2.12 & 7.93 & 19.23 & 11.30 & 11.77 & 13.53 & 1.76 & 8.09 & 8.97 & 0.88 & 26.62 & 6.24 & -20.38 & 12.64 & 11.66 & -0.98 & 8.40 & 13.62 & 5.22 \\ \midrule

 \textbf{mean} & - & - & 2.39 & - & - & 12.65 & - & - & 0.94 & - & - & -2.70 & - & - & -20.27 & - & - & 1.19 & - & - & 11.94\\

 \textbf{std dev} & - & - & 1.23 & - & - & 2.53 & - & - & 1.96 & - & - & 1.83 & - & - & 5.17 & - & - & 2.17 & - & - & 4.81 \\ \bottomrule 
\end{tabular}
}
\vspace{1pt}\caption{\textbf{Compositional Integration L2 Object-Attribute Accuracy (Raw Results).} We report the difference between decomposed ($D$) and composed ($C$) accuracy per complexity level, along with the mean and standard deviation of $\Delta$ for each model in OA setting. Positive $\Delta$ values indicate that model performs better independently than jointly, highlighting compositional binding difficulty.}
\label{tab:ci_l2_oa}
\end{table}
The results in Table \ref{tab:ci_l2_oa} indicate a clear integration gap with positive $\Delta$ values for most models, where the additional cost of the joint compositional reasoning consistently dominates the pure cost of the independent primitive recognition. In contrast, both NegCLIP and CE-CLIP serve as notable exceptions to that observation. Both models consistently display negative $\Delta$ values, which is a clear indicator of superior performance on composed tasks compared to their decomposed counterparts. This behavior is more strongly demonstrated in CE-CLIP, with a relatively low mean of $-20.27$ for$\Delta$. These two exceptional cases might imply that through specialized training objectives such as compositional binding or contrastive discrimination of hard negatives, the model's behavior in composed and decomposed settings can be altered.

\begin{table}[h] 
\centering 
\scriptsize 
\setlength{\tabcolsep}{2.2pt} 
\resizebox{\columnwidth}{!}{
\begin{tabular}{@{}c|ccc|ccc|ccc|ccc|ccc|ccc|ccc@{}} 
\toprule 

 & \multicolumn{3}{c|}{\textbf{OpenCLIP}} & 
\multicolumn{3}{c|}{\textbf{SigLIP 2}} & 
\multicolumn{3}{c|}{\textbf{PE-CLIP}} & 
\multicolumn{3}{c|}{\textbf{NegCLIP}} & 
\multicolumn{3}{c|}{\textbf{CE-CLIP}} & 
\multicolumn{3}{c}{\textbf{BLIP-L}} & 
\multicolumn{3}{c}{\textbf{Qwen3}} \\ 

 \textbf{Complexity Level} & \textbf{C} & \textbf{D} & 
\textbf{$\Delta$} & \textbf{C} & \textbf{D} & 
\textbf{$\Delta$} & \textbf{C} & \textbf{D} & 
\textbf{$\Delta$} & \textbf{C} & \textbf{D} & 
\textbf{$\Delta$} & \textbf{C} & \textbf{D} & 
\textbf{$\Delta$} & \textbf{C} & \textbf{D} & 
\textbf{$\Delta$} & \textbf{C} & \textbf{D} & 
\textbf{$\Delta$} \\ \midrule 

 3 & 28.79 & 41.34 & 12.54 & 18.89 & 33.73 & 14.84 & 34.71 & 36.30 & 1.59 & 41.46 & 46.17 & 4.71 & 49.90 & 38.49 & -11.41 & 44.23 & 39.55 & -4.69 & 33.11 & 45.4 & 12.29 \\ 

 4 & 21.39 & 37.81 & 16.42 & 12.71 & 28.69 & 15.99 & 28.00 & 30.60 & 2.61 & 35.01 & 39.12 & 4.11 & 41.24 & 31.86 & -9.38 & 29.07 & 34.04 & 4.97 & 24.61 & 40.46 &	15.84  \\ 

 5 & 17.80 & 34.75 & 16.95 & 9.21 & 24.66 & 15.46 & 22.85 & 26.94 & 4.09 & 27.79 & 33.14 & 5.35 & 33.55 & 23.58 & -9.96 & 22.68 & 28.70 & 6.02 & 20.68 & 33.55 &	12.87 \\ 

 6 & 10.13 & 28.03 & 17.89 & 5.85 & 14.68 & 8.83 & 14.53 & 18.07 & 3.53 & 19.63 & 25.19 & 5.56 & 23.89 & 18.36 & -5.53 & 16.97 & 19.77 & 2.80 & 16.86 & 24.46 & 7.59 \\ 

 7 & 7.50 & 26.42 & 18.92 & 4.76 & 13.16 & 8.40 & 13.10 & 16.02 & 2.92 & 15.12 & 21.87 & 6.75 & 19.94 & 15.62 & -4.32 & 13.22 & 17.58 & 4.36 & 13.56 & 22.98 & 9.41 \\ 

 8 & 5.93 & 23.93 & 18.01 & 4.20 & 10.26 & 6.06 & 9.44 & 11.89 & 2.44 & 10.75 & 19.11 & 8.37 & 15.17 & 12.50 & -2.67 & 11.04 & 13.74 & 2.70 & 12.94 & 18.83 & 5.88 \\ 

 9 & 4.31 & 22.68 & 18.37 & 2.65 & 8.40 & 5.75 & 6.71 & 9.17 & 2.46 & 8.69 & 15.21 & 6.52 & 11.98 & 11.05 & -0.93 & 8.72 & 10.70 & 1.98 & 10.91 & 14.72 & 3.81 \\ 

 10 & 4.17 & 22.46 & 18.29 & 2.40 & 7.31 & 4.91 & 6.24 & 8.44 & 2.21 & 6.68 & 12.88 & 6.21 & 10.58 & 8.51 & -2.07 & 7.01 & 9.35 & 2.34 & 10.45 & 13.15 & 2.69 \\ 

 11 & 2.30 & 21.01 & 18.72 & 2.16 & 5.31 & 3.15 & 4.29 & 6.99 & 2.71 & 6.96 & 10.87 & 3.91 & 8.36 & 7.37 & -0.99 & 6.68 & 7.54 & 0.86 & 9.38	& 10.97 & 1.59 \\ 

 12 & 2.45 & 21.94 & 19.49 & 1.54 & 4.63 & 3.09 & 2.91 & 5.26 & 2.35 & 5.78 & 9.78 & 4.00 & 8.27 & 5.89 & -2.38 & 4.77 & 6.41 & 1.65 & 8.33 & 9.46 & 1.13 \\ \midrule 

 \textbf{mean} & - & - & 17.56 & - & - & 8.64 & - & - & 2.69 & - & - & 5.54 & - & - & -4.96 & - & - & 2.30 & - & - & 7.31 \\ 

 \textbf{std dev} & - & - & 1.98 & - & - & 5.04 & - & - & 0.70 & - & - & 1.44 & - & - & 3.93 & - & - & 2.93 & - & - & 5.16 \\ \bottomrule 
\end{tabular}
}
\vspace{1pt}\caption{\textbf{Compositional Integration L2 Object-Relation Accuracy (Raw Results).} We report the difference between decomposed ($D$) and composed ($C$) accuracy per complexity level, along with the mean and standard deviation of $\Delta$ for each model in OR setting. Positive $\Delta$ values indicate that model performs better independently than jointly, highlighting compositional binding difficulty.}
\label{tab:ci_l2_or}
\end{table}
The Object-Relation results in \ref{tab:ci_l2_or}
follow the trend of considerable integration penalty across different models, with the most severe performance drops in OpenCLIP and SigLIPv2 models. Remarkably, in this context, NegCLIP does not retain the outlier position it has in the Object-Attribute setting, shifting to positive $\Delta$ values across all complexity levels. Conversely, CE-CLIP continues to be a counterpoint in this setting as well, with a negative mean $\Delta$ of $-4.96$. This suggests that CE-CLIP's contrastive intra-modal and cross-modal training objectives might effectively mitigate the primary integration burden in joint compositional reasoning. 

\begin{table}[h] 
\centering 
\scriptsize 
\setlength{\tabcolsep}{2.2pt} 
\resizebox{\columnwidth}{!}{
\begin{tabular}{@{}c|ccc|ccc|ccc|ccc|ccc|ccc|ccc@{}} 
\toprule 

 & \multicolumn{3}{c|}{\textbf{OpenCLIP}} & 
\multicolumn{3}{c|}{\textbf{SigLIP 2}} & 
\multicolumn{3}{c|}{\textbf{PE-CLIP}} & 
\multicolumn{3}{c|}{\textbf{NegCLIP}} & 
\multicolumn{3}{c|}{\textbf{CE-CLIP}} & 
\multicolumn{3}{c}{\textbf{BLIP-L}} & 
\multicolumn{3}{c}{\textbf{Qwen3}} \\ 

 \textbf{Complexity Level} & \textbf{C} & \textbf{D} & 
\textbf{$\Delta$} & \textbf{C} & \textbf{D} & 
\textbf{$\Delta$} & \textbf{C} & \textbf{D} & 
\textbf{$\Delta$} & \textbf{C} & \textbf{D} & 
\textbf{$\Delta$} & \textbf{C} & \textbf{D} & 
\textbf{$\Delta$} & \textbf{C} & \textbf{D} & 
\textbf{$\Delta$} & \textbf{C} & \textbf{D} & 
\textbf{$\Delta$} \\ \midrule 

 4 & 20.54 & 19.85 & -0.69 & 18.99 & 25.48 & 6.48 & 25.56 & 28.06 & 2.49 & 33.78 & 36.66 & 2.88 & 41.03 & 31.15 & -9.88 & 33.47 & 33.08 & -0.38 & 25.1 & 35.98 &	10.87\\ 

 5 & 16.83 & 18.34 & 1.51 & 17.60 & 22.05 & 4.44 & 23.79 & 26.31 & 2.52 & 27.14 & 29.94 & 2.79 & 34.80 & 23.10 & -11.69 & 24.85 & 27.14 & 2.29 & 19.44 & 31.51 & 12.06\\ 

 6 & 10.53 & 12.31 & 1.78 & 12.98 & 15.63 & 2.64 & 16.69 & 18.80 & 2.11 & 20.77 & 24.86 & 4.08 & 25.06 & 17.21 & -7.84 & 19.19 & 20.68 & 1.49 & 16.21 & 23.96 &
 7.74\\ 

 7 & 7.72 & 8.59 & 0.86 & 9.11 & 11.46 & 2.35 & 11.51 & 13.29 & 1.77 & 15.06 & 19.52 & 4.46 & 20.10 & 15.45 & -4.65 & 14.58 & 16.69 & 2.11 & 14.21 & 21.36 & 7.15\\ 

 8 & 5.73 & 6.52 & 0.79 & 6.71 & 8.69 & 1.97 & 9.04 & 11.70 & 2.66 & 12.05 & 16.05 & 4.00 & 16.79 & 12.64 & -4.15 & 11.16 & 14.03 & 2.86 & 12.7	& 17.79	& 5.09\\ 

 9 & 4.51 & 4.97 & 0.46 & 5.69 & 6.73 & 1.03 & 6.78 & 9.01 & 2.22 & 8.64 & 13.10 & 4.45 & 13.77 & 8.49 & -5.28 & 9.32 & 9.63 & 0.31 & 10.67 & 13.57 & 2.9 \\ 

 10 & 2.92 & 3.80 & 0.89 & 3.96 & 6.20 & 2.24 & 5.84 & 7.04 & 1.19 & 7.35 & 10.53 & 3.18 & 11.50 & 7.92 & -3.59 & 6.99 & 8.81 & 1.82 & 10.33 & 10.54 & 0.2\\ 

 11 & 2.36 & 3.11 & 0.75 & 3.21 & 5.03 & 1.82 & 4.28 & 6.05 & 1.76 & 5.08 & 9.05 & 3.96 & 8.24 & 6.80 & -1.44 & 6.48 & 7.12 & 0.64 & 7.92	& 9.9 & 1.98\\ 

 12 & 2.07 & 3.15 & 1.08 & 2.99 & 3.53 & 0.54 & 4.35 & 5.32 & 0.97 & 4.45 & 8.37 & 3.91 & 7.39 & 5.00 & -2.39 & 5.05 & 6.36 & 1.30 & 8.59 & 8.64 & 0.05\\ \midrule 

 \textbf{mean} & - & - & 0.83 & - & - & 2.61 & - & - & 1.97 & - & - & 3.75 & - & - & -5.66 & - & - & 1.38 & - & - & 5.34\\ 

 \textbf{std dev} & - & - & 0.70 & - & - & 1.82 & - & - & 0.59 & - & - & 0.64 & - & - & 3.45 & - & - & 1.03 & - & - & 4.42\\ \bottomrule 
\end{tabular} 
}
\vspace{1pt}
\caption{\textbf{Compositional Integration L3 Object-Attribute-Relation (Raw Results).} We report the difference between decomposed ($D$) and composed ($C$) accuracy per complexity level, along with the mean and standard deviation of $\Delta$ for each model in OAR setting. Positive $\Delta$ values indicate that model performs better independently than jointly, highlighting compositional binding difficulty.}
\label{tab:ci_l3}
\end{table}
In most structurally complex scene composition (OAR) settings, the results in Table \ref{tab:ci_l3} show a considerable integration overhead, despite lower discrepancies between composed and decomposed settings with lower $\Delta$ than in simpler OR and OA settings. Interestingly, NegCLIP showed the largest gap in this setting, with a mean $\Delta$ of $3.75$, reinforcing the observations in the OR and OA settings that hard-negative discrimination with attributes does not directly translate to whole-scene compositions. In contrast, CE-CLIP maintains its position as a robust outlier, with a much larger gap in the opposite direction, a mean $\Delta$ of $-5.66$. This further signifies CE-CLIP's effectiveness in reinforcing compositional integration. 

\section{Compute Resources}
\label{app:compute}

Caption generation and hard negative generation were performed using the GPT-4o-mini API~\cite{gpt4o-mini}. All model evaluations and perplexity computations were run on a single NVIDIA A100 GPU. Post-processing and regression analysis were performed on CPU with 4 parallel workers. Among the evaluated models, all contrastive models (OpenCLIP, SigLIPv2, NegCLIP, PE-CLIP) and BLIP completed evaluation within 2--3 hours per model. Qwen3-VL-Embedding-8B required approximately 8 
hours due to its larger model size.

\begin{figure*}[t]
    \centering
    \begin{subfigure}{0.8\textwidth}
        \centering
        \includegraphics[width=\textwidth, trim={3.2cm 4cm 4.3cm 0.5cm},clip]{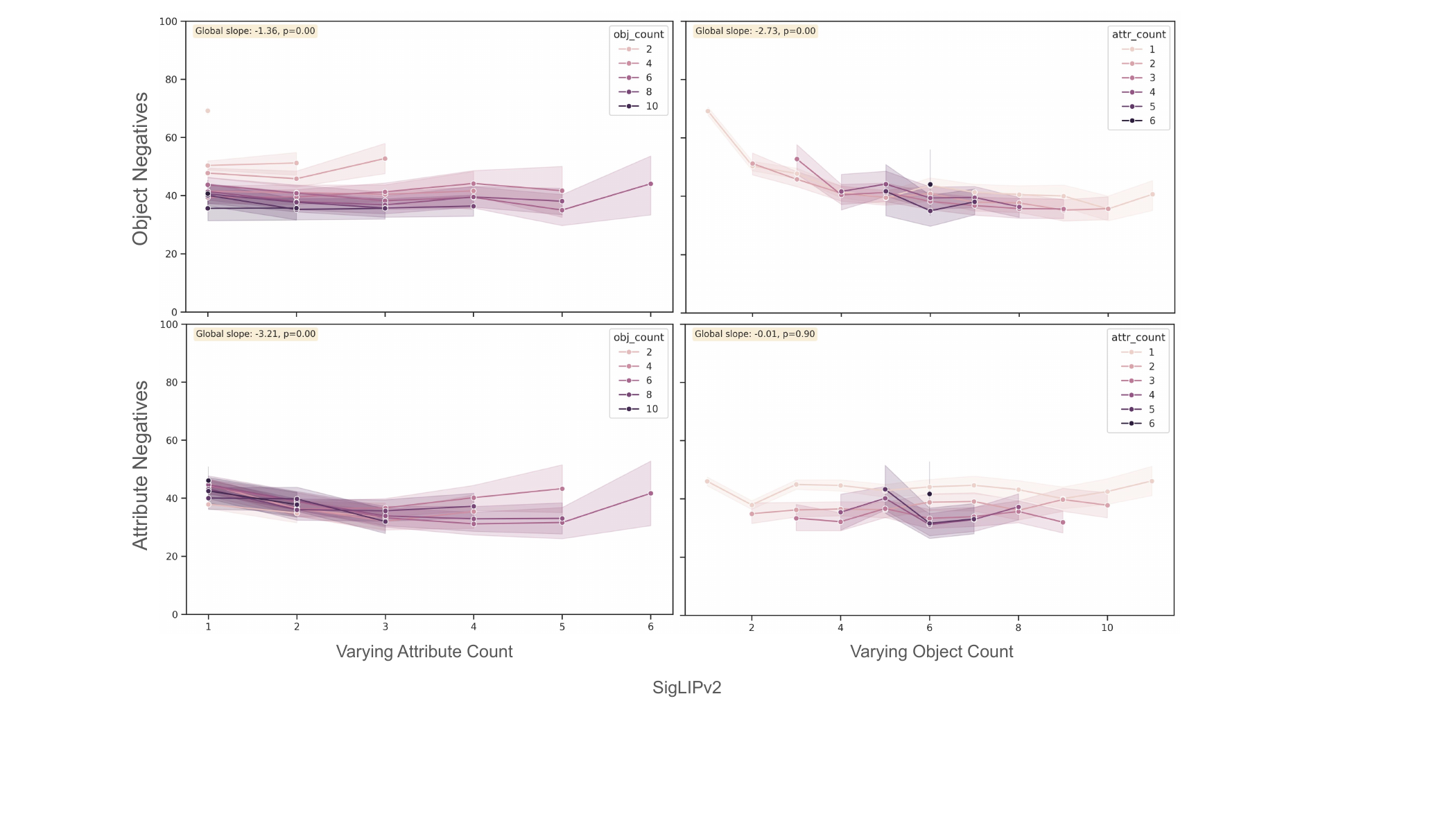}
        \caption{SigLIPv2}
    \end{subfigure}
    
    \vspace{0.7em}
    
    \begin{subfigure}{0.8\textwidth}
        \centering
        \includegraphics[width=\textwidth, trim={3.2cm 4cm 4.3cm 0.5cm},clip]{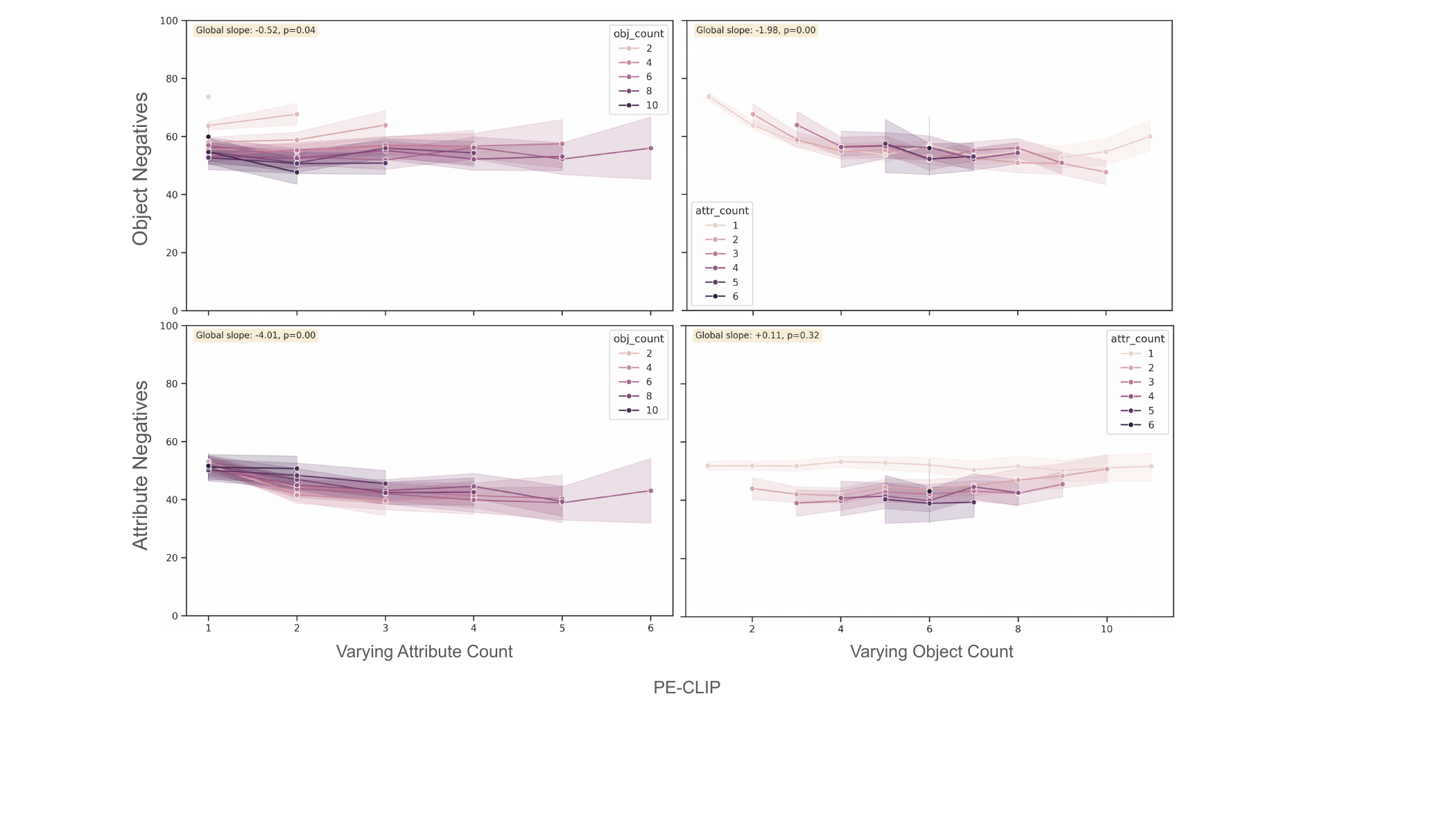}
        \caption{PE-CLIP}
    \end{subfigure}
    
    \caption{Skill load analysis on L2 (OA) captions for SigLIPv2 and 
PE-CLIP. Each model shows four panels: object negatives (top) and 
attribute negatives (bottom), each varying attribute count (left) and 
object count (right) while holding the other fixed.}
    \label{fig:skill_load_oa_1}
\end{figure*}

\begin{figure*}[t]
    \centering
    \begin{subfigure}{0.8\textwidth}
        \centering
        \includegraphics[width=\textwidth, trim={3.2cm 4cm 4.3cm 0.5cm},clip]{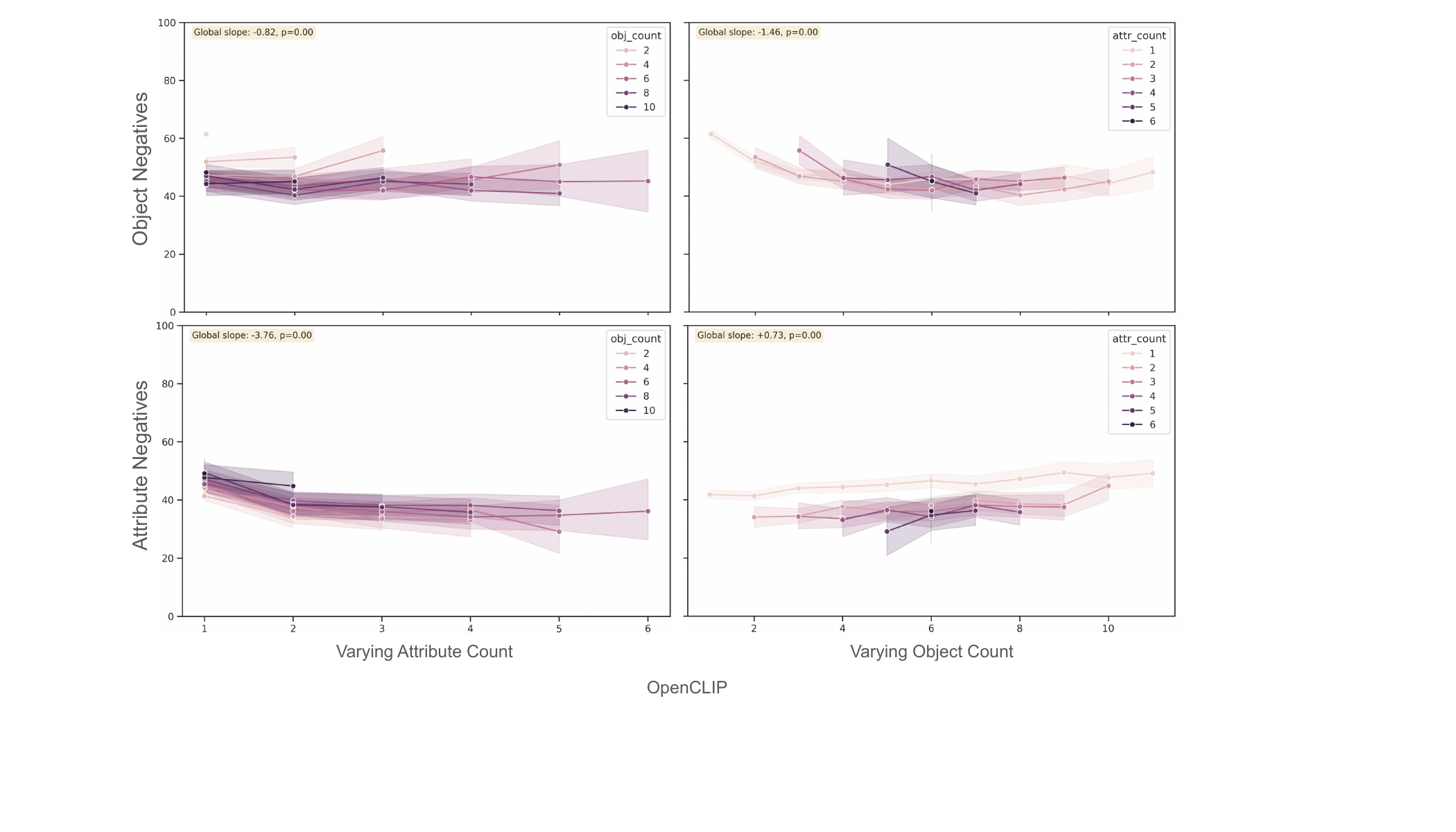}
        \caption{OpenCLIP}
    \end{subfigure}
    
    \vspace{0.7em}
    
    \begin{subfigure}{0.8\textwidth}
        \centering
        \includegraphics[width=\textwidth, trim={3.2cm 4cm 4.3cm 0.5cm},clip]{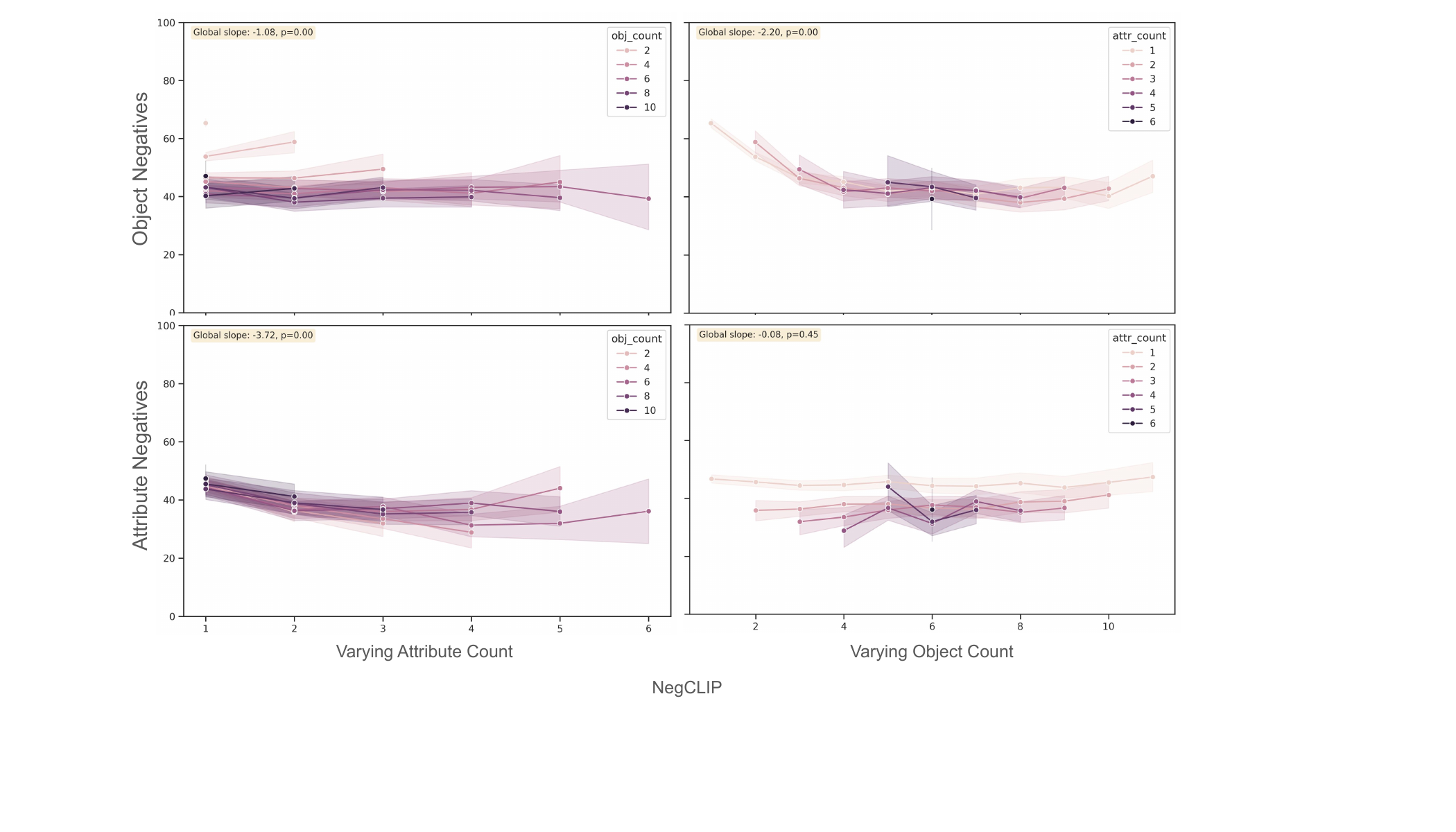}
        \caption{NegCLIP}
    \end{subfigure}
    
    \vspace{0.7em}
    
    \begin{subfigure}{0.8\textwidth}
        \centering
        \includegraphics[width=\textwidth, trim={3.2cm 4cm 4.3cm 0.5cm},clip]{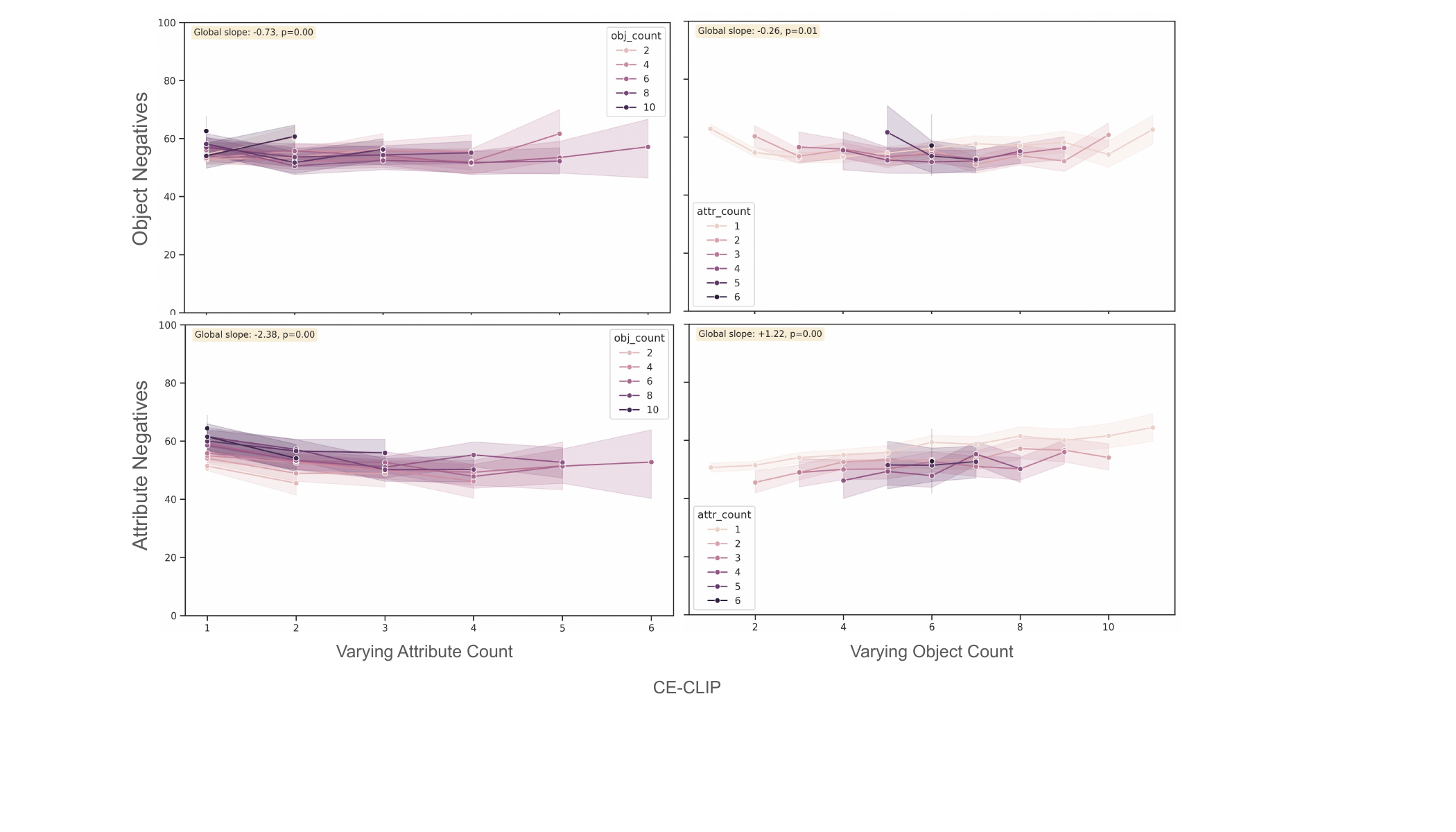}
        \caption{CE-CLIP}
    \end{subfigure}
    
    \caption{Skill load analysis on L2 (OA) captions for OpenCLIP, NegCLIP, 
and CE-CLIP. Layout follows Figure~\ref{fig:skill_load_oa_1}.}
    \label{fig:skill_load_oa_2}
\end{figure*}

\begin{figure*}[t]
    \centering
    \begin{subfigure}{0.8\textwidth}
        \centering
        \includegraphics[width=\textwidth, trim={3.2cm 4cm 4.3cm 0.5cm},clip]{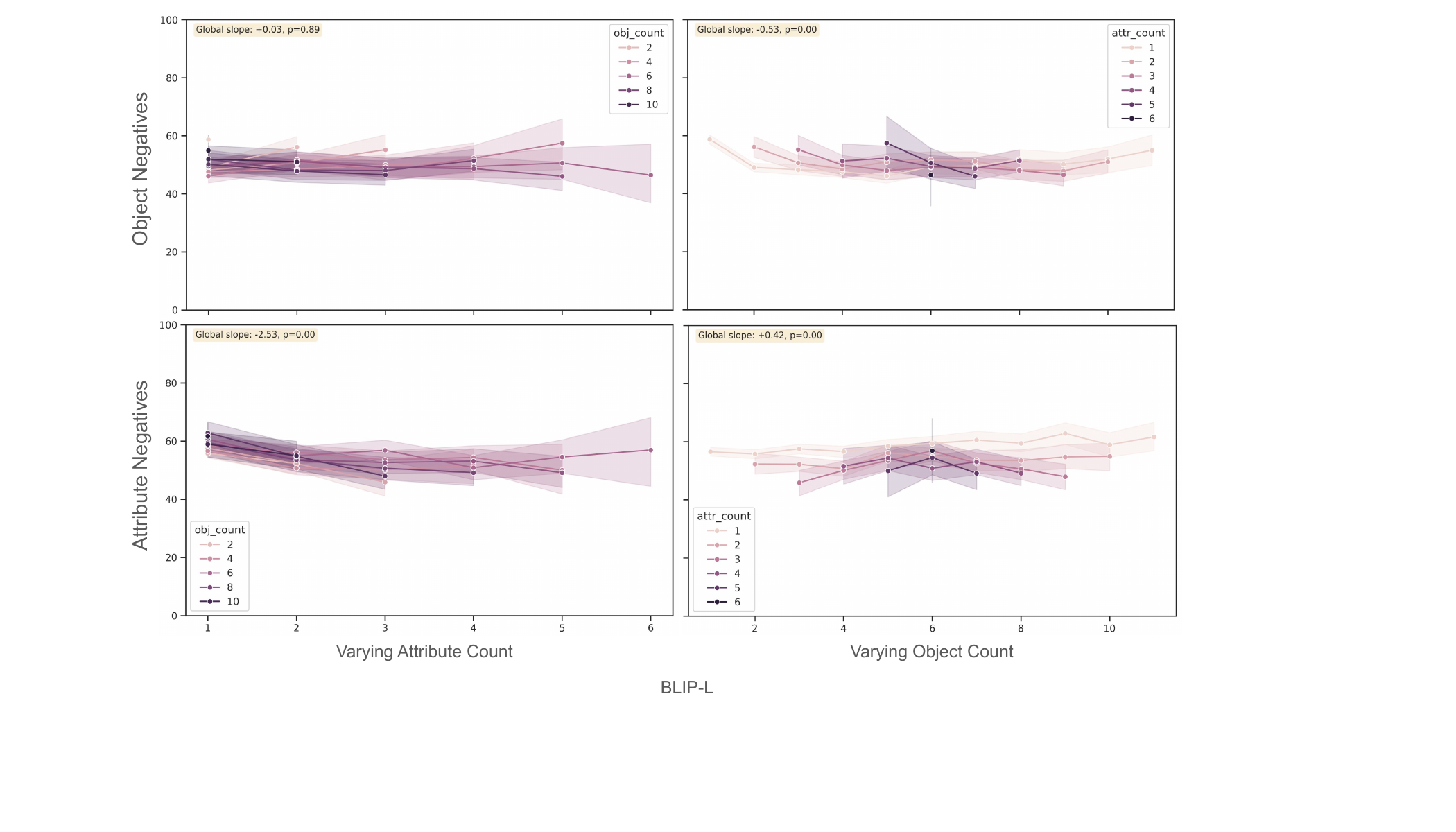}
        \caption{BLIP-L}
    \end{subfigure}
    
    \vspace{0.7em}
    
    \begin{subfigure}{0.8\textwidth}
        \centering
        \includegraphics[width=\textwidth, trim={3.2cm 4cm 4.3cm 0.5cm},clip]{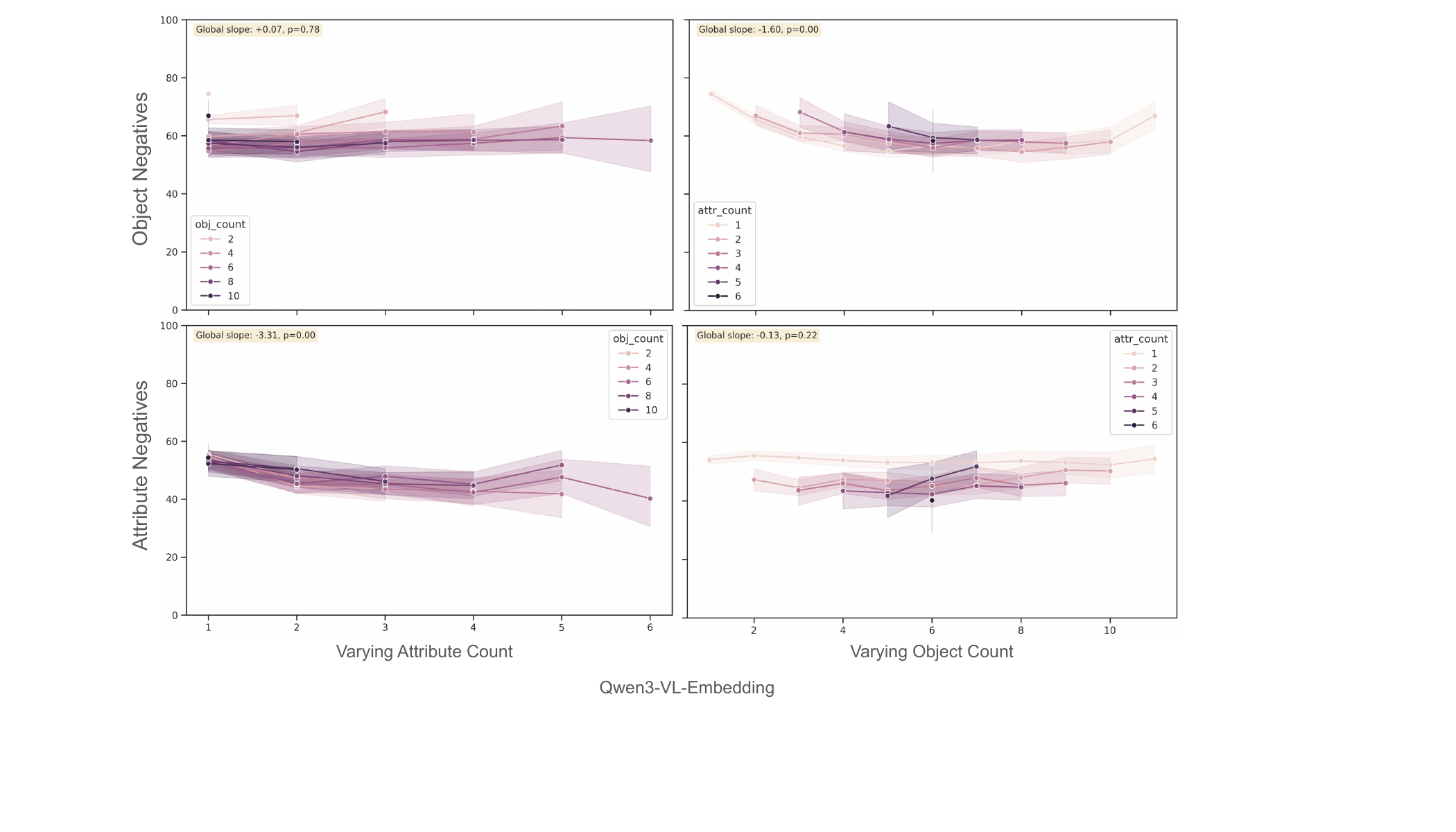}
        \caption{Qwen3-VL-Embedding}
    \end{subfigure}
    
    \caption{Skill load analysis on L2 (OA) captions for BLIP and 
Qwen3-VL-Embedding. Layout follows Figure~\ref{fig:skill_load_oa_1}.}
    \label{fig:skill_load_oa_3}
\end{figure*}

\section{Skill Load: Visualization}
\label{app:skill_load_viz}

To complement the regression coefficients reported in 
Table~\ref{tab:skill_load}, we visualize the underlying degradation 
patterns directly. For each model, we plot R@1 accuracy as a function 
of one primitive count while holding the other fixed, with separate lines 
for each value of the controlled count. Each figure shows four panels per 
model: object negatives (top) and attribute negatives (bottom), each 
varying attribute count (left) and object count (right). The global slope 
reported in each panel corresponds to the regression coefficient in 
Table~\ref{tab:skill_load}, estimated via OLS while controlling for all 
primitive counts.

Figures~\ref{fig:skill_load_oa_1}--\ref{fig:skill_load_oa_3} show 
results for L2 (OA) captions across all seven models. The pattern is 
consistent throughout: for object negatives, varying the attribute count 
produces flat lines while varying the object count produces consistently 
declining lines. For attribute negatives, varying the object count 
produces flat or inconsistent lines while varying the attribute count 
produces steep and consistent declines. This directly illustrates the 
self-load dominance reported in Table~\ref{tab:skill_load}, each skill 
degrades primarily under the weight of its own primitive count, with 
cross-load effects remaining small and inconsistent across all models 
and architectures.



\end{document}